\documentclass[10pt,twocolumn,letterpaper]{article}

\usepackage[pagenumbers]{cvpr} 
\usepackage[accsupp]{axessibility}
\usepackage{graphicx}
\usepackage{amsmath}
\usepackage{amssymb}
\usepackage{booktabs}

\usepackage[pagebackref,breaklinks,colorlinks]{hyperref}

\usepackage[capitalize]{cleveref}
\crefname{section}{Sec.}{Secs.}
\Crefname{section}{Section}{Sections}
\Crefname{table}{Table}{Tables}
\crefname{table}{Tab.}{Tabs.}

\def\cvprPaperID{4055} 
\def\confName{CVPR}
\def\confYear{2022}

\newif\ifdraft
\drafttrue

\newcommand{\R}{\mathbb{R}}

\newcommand{\comment}[1]{}

\newcommand{\parag}[1]{{\bf{#1}}}

\newcommand{\vp}{\mathbf{p}}

\newcommand{\vt}{\mathbf{t}}
\newcommand{\vu}{\mathbf{u}}
\newcommand{\vv}{\mathbf{v}}
\newcommand{\vw}{\mathbf{w}}
\newcommand{\vx}{\mathbf{x}}

\newcommand{\vz}{\mathbf{z}}

\newcommand{\mB}{\mathbf{B}}

\newcommand{\mF}{\mathbf{F}}

\newcommand{\mH}{\mathbf{H}}
\newcommand{\mI}{\mathbf{I}}

\newcommand{\mM}{\mathbf{M}}

\newcommand{\mW}{\mathbf{W}}
\newcommand{\mX}{\mathbf{X}}

\newcommand{\cD}{\mathcal D}

\newcommand{\cG}{\mathcal G}

\newcommand{\cL}{\mathcal L}

\newcommand{\cN}{\mathcal N}

\begin{document}

\title{GANSeg: Learning to Segment by Unsupervised Hierarchical Image Generation}

\author{Xingzhe He \hspace{5mm} Bastian Wandt \hspace{5mm} Helge Rhodin \\
University of British Columbia\\
{\tt\small \{xingzhe, wandt, rhodin\}@cs.ubc.ca}}
\maketitle

\begin{abstract}
Segmenting an image into its parts is a common preprocess for high-level vision tasks such as image editing.
However, annotating masks for supervised training is expensive. Weakly-supervised and unsupervised methods exist, but they depend on the comparison of pairs of images, such as from multi-views, frames of videos, and image augmentation,
which limit their applicability.
To address this, we propose a GAN-based approach that generates images conditioned on \emph{latent} masks, thereby alleviating full or weak annotations required by previous approaches.
We show that such mask-conditioned image generation can be learned faithfully when conditioning the masks in a hierarchical manner on 2D \emph{latent points} that define the position of parts explicitly.
Without requiring supervision of masks or points, this strategy increases robustness of mask to viewpoint and object position changes.
It also lets us generate image-mask pairs for training a segmentation network, which outperforms state-of-the-art unsupervised segmentation methods on established benchmarks. 
Code can be found at \href{https://github.com/xingzhehe/GANSeg}{https://github.com/xingzhehe/GANSeg}.
\end{abstract}

\section{Introduction}

\begin{figure*}[t]
\begin{center}
   \includegraphics[width=0.9\linewidth]{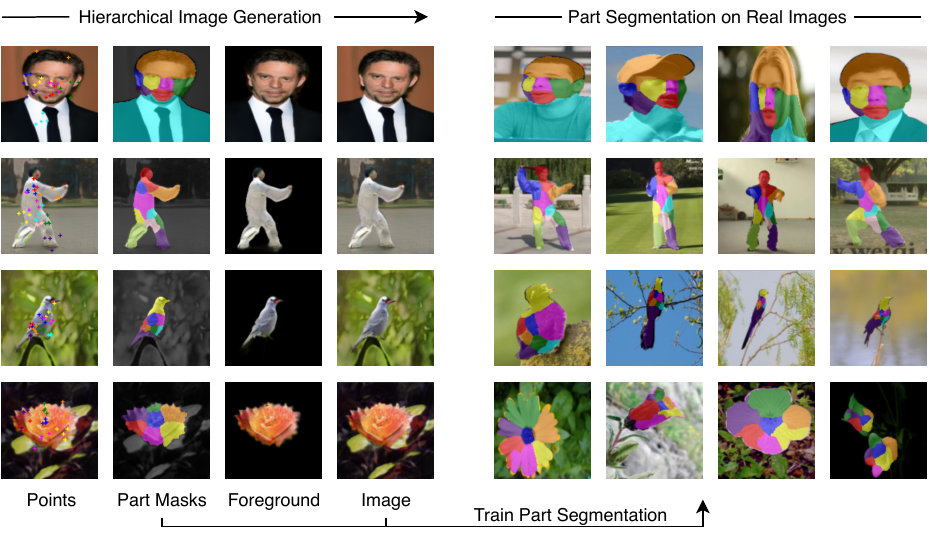}
\end{center}
\vspace{-18pt}
   \caption{\textbf{GANSeg.} A segmentation network (right) is trained on mask-image pairs generated by a new hierarchical image generator (left; from points, to mask, over foreground to the image). It is unsupervised and applies to  faces, persons, birds, and flowers.}
\label{fig:teaser}
\end{figure*}

This paper tackles the problem of unsupervised part segmentation.
Discovering object parts in images is a fundamental problem in computer vision as parts provide an intermediate representation that is robust to object appearance and pose variation \cite{hung2019scops, siarohin2021motion}. Many high-level tasks benefit from part representations, such as 3D reconstruction \cite{li2020self, zuffi2015}, pose estimation \cite{Guler2018DensePose, kiefel2014}, and image editing \cite{he2021latentkeypointgan, zhu2020sean}. Keypoints and part segmentation maps are among the most commonly used forms. However, their supervised training \cite{wei2016convolutional, ronneberger2015u, he2017mask} requires pixel-level annotations for every new application domain since labels hardly generalize to other object categories and the number of parts and their granularity vary across tasks.

On the side of keypoint detection, several unsupervised detectors exist \cite{zhang2018unsupervised, jakab2018unsupervised} but segmentation methods are still in their infancy \cite{hung2019scops, liu2021unsupervised}.
%
Segmenting parts without pixel-level annotation is  difficult because it requires 
disentangling parts from other parts and the foreground from the background. Existing unsupervised\footnote{Most existing literature refers to \emph{unsupervised} when training on single images without annotation, and \emph{self-supervised} when training on auxiliary tasks using multi-views or videos. We follow this convention.} methods mainly follow the same strategy as applied for unsupervised keypoint detection \cite{thewlis2017unsupervised}. Real images are  transformed by an affine map or a thin plate spline to find those parts that are equivariant under the known deformation.
For precise reconstruction they require additional information, such as saliency maps \cite{hung2019scops} or assume the objects to be consistently centered \cite{liu2021unsupervised}, which is constraining. For example, when applied to face datasets \cite{liu2015faceattributes}, the neck and shoulders are often ignored although part of almost every image. 


Our goal is to improve the unsupervised part segmentation task.
We propose to first train a generative adversarial network (GAN) \cite{goodfellow2014generative} to generate images that are internally conditioned on \emph{latent masks}. This GAN formulation alleviates the dependency on image pairs and pre-defined image transformation in existing autoencoder networks. In this way, the network learns the part distribution from the dataset instead of from pre-defined image transformations. Subsequently, we use the generator to synthesize virtually infinite mask-image pairs for training a segmentation network. Figure~\ref{fig:teaser} provides an overview of our model.

The key question we address is how to design a GAN that generates images with part segmentation masks that are meaningful, 
i.e., group pixels into regions that typically move together and have shared appearance across images.
We start from a backbone architecture that is borrowed from supervised segmentation networks \cite{ronneberger2015u, chen2017rethinking} and the GAN strategy is inspired by its recent application to unsupervised keypoint detection \cite{he2021latentkeypointgan}. 
Our innovation is the hierarchical generation of the image via multiple abstraction levels, including the use of masks. 
In the first level, we use Gaussian noise to generate part appearance embeddings and a set of \emph{2D latent points}. Unlike \cite{he2021latentkeypointgan} which continues straight from points to image generation, we first group points to define the position and the scale of each part.
In the second abstraction level, we use a part-relative positional encoding to generate 2D feature maps and then generate the mask with a CNN. 
In the third level, the foreground image is generated from a combination of feature maps with corresponding appearance embeddings.
Independently, a background image is generated with a randomized position to disentangle foreground and background.
Finally the foreground is blended with the background. The generated masks are used here a second time to define the blend weight.

The key to the success of our GAN framework are several design choices that preserve translational equivariance of parts \cite{karras2021alias}, which are not applicable to the traditional autoencoder approaches, as explained in Appendix~\ref{sec:theory}. As a result, by not knowing the absolute location, the convolutional network is forced to condition the image purely on the spatial extent of the masks; moving the part mask will move the image part. This is a crucial inductive bias in our unsupervised learning.
%
%
Our contributions are threefold:
\begin{enumerate}
    \item 
    An unsupervised GAN approach to generate mask-image pairs for training part segmentation;
    \item A novel hierarchical image generator, which encourages the segmentation of meaningful parts;
    \item Alleviating prior assumptions on saliency maps and object position.
\end{enumerate}

\textbf{Ethics - Risks.} GANs can be abused for creating deep fakes. However, our method does not work towards editing nor improving image quality but scene understanding. Our final output is a detector, which can not be abused to generate new images but unwanted surveillance applications are a risk.
\textbf{Benefits.} Since our method is entirely unsupervised, it could be applied on objects, animals, or situations that have not yet been labeled. 

\section{Related Work}
\noindent\parag{Unsupervised Landmark Detection} methods discover keypoints in the images without any supervision signals. Most existing works discover keypoints by comparing pairs of images of the same object category. The common idea is that the keypoints change as the image changes.  The change can be inferred from videos \cite{siarohin2019animating, kulkarni2019unsupervised, minderer2019unsupervised, dong2018supervision, kim2019unsupervised, jakab2020self} and multi-views \cite{suwajanakorn2018discovery, rhodin2018unsupervised, Rhodin_2019_CVPR} of the same object category.  While videos and multi-views naturally contain pairs of images, those relying on unsupervised learning from image collections require pairs created by pre-defined random transformations \cite{thewlis2017unsupervised, zhang2018unsupervised, jakab2018unsupervised, lorenz2019unsupervised} and learned transformation \cite{wu2019transgaga, xu2020unsupervised} that are tuned for the dataset. Their underlying idea is similar. Keypoints must follow the transformation that is applied to the original images---equivariance. Recently, \cite{he2021latentkeypointgan} introduced an alternative. They use a GAN to generate images along with corresponding latent keypoints and use them to train a detector. By contrast to sparse keypoints, our goal is to generate masks at the pixel level, which we attain by introducing a hierarchical generator.

\noindent\parag{Unsupervised Foreground Segmentation} aims to segment foreground objects from the background in an unsupervised manner. 
Gupta et al. \cite{gupta2020patchvae} learn a patch level mask of foreground to benefit representation learning, which, however, is only coarse. Singh et al. \cite{singh2019finegan} use a multi-stage GAN to disentangle the foreground shape and texture but they only focus on editing.
Bielski et al.~\cite{bielski2019emergence} propose to re-position the generated foreground to disentangle it from the background. 
Yang et al.~\cite{yang2019unsupervised} propose
 Contextual Information Separation (CIS),
a general objective to segment the foreground by maximizing the error of inpainting the mask and its complement. It was first applied to optical flow and subsequently to color images by \cite{savarese2021information,yang2021dystab, katircioglu2021self,katircioglu2021human}.
Chen et al.~\cite{chen2019unsupervised} achieved unsupervised foreground segmentation by resampling the foreground appearance to disentangle foreground and background. Voynov et al.~\cite{voynov2021object} and Yang et al.~\cite{yang2021unsupervised} introduce comparison-based segmentation methods using pre-trained GANs, achieving better results than generation-based methods. 
As pointed out by \cite{bielski2019emergence, chen2019unsupervised}, such unsupervised methods can easily get into trivial solutions, where the background 
includes the whole foreground.
To counteract, we introduce two losses that overcome
trivial solutions. Besides, we found that preserving translation equivariance in the network architecture can naturally mitigate trivial solutions. This is another important reason why we chose to use a GAN, as we will explain in Section~\ref{sec:method} and supplementary. In comparison to the above methods, this yields comparable results even if they are specialized for foreground separation while we provide more fine-grained part segmentation.

\noindent\parag{Unsupervised Part Segmentation} aims at pixel-level masks for multiple parts of an object, including
separating foreground from background without mask annotation.
Collins et al. \cite{collins2018deep} use matrix factorization to find similar parts in images, but it requires test images at training time, which makes it computational prohibitive. Hung et al. \cite{hung2019scops} draws lessons from unsupervised keypoint detection and extends them to predict the part segmentation masks of an object using various loss functions that
preserve the geometry and semantic consistency of the masks. However, it needs off-the-shelf saliency maps or ground truth background masks. Liu et al. \cite{liu2021unsupervised} alleviate the need for the background mask, but they use a central prior to constrain the object mask to the center of the image, which can be a constraining bias. For example, in portrait images, the hair is often not masked. Temporal information can be used \cite{siarohin2021motion, gao2021unsupervised} to achieve better segmentation results. In comparison to all of these approaches, our model uses less information (single images without video or saliency map) yet outperform these on half of the most established metrics and datasets, as we evaluate in our  experiments, Section~\ref{experiment}. The recently proposed DatasetGAN \cite{zhang2021datasetgan}
also utilize the a GAN to segment parts, but they still need some ground truth masks as parts are not explicitly disentangled within the network.

\section{Method} \label{sec:method}
\begin{figure}[t]
\begin{center}
   \includegraphics[width=0.98\linewidth]{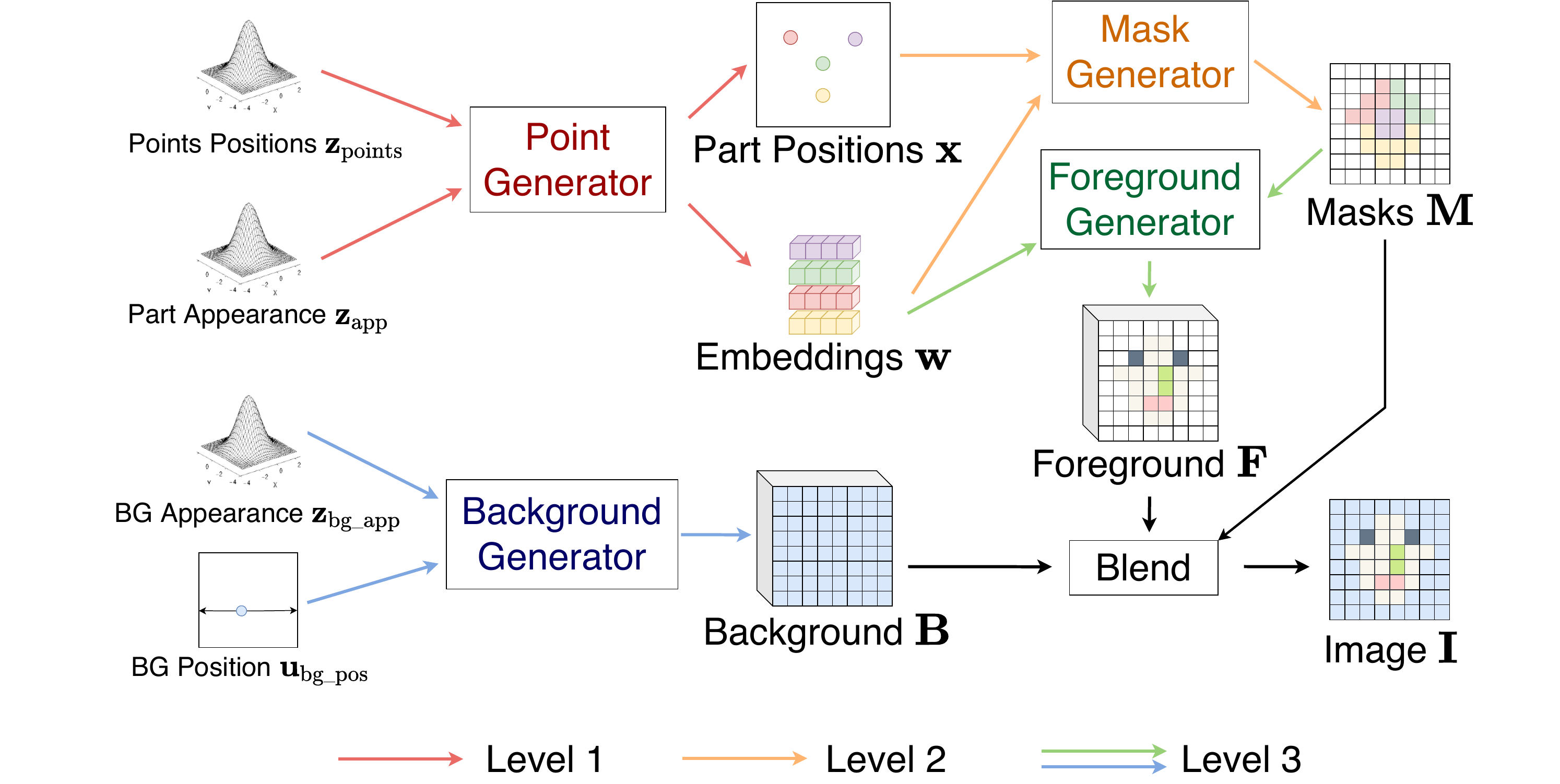}
\end{center}
   \caption{\textbf{Hierarchical generation overview.} \textbf{Level 1 (red):} The Point Generator converts two Gaussian noise vectors to part position and appearance embedding. \textbf{Level 2 (orange):} The Mask Generator turns part positions and embeddings into masks defining the part support. \textbf{Level 3 (green and blue):} The Foreground Generator uses the quantities from the previous level to generate a foreground image that is finally blended with the independently generated background.}
\label{fig:overview}
\end{figure}

We train a Generative Adversarial Network (GAN) \cite{goodfellow2014generative} to generate points, part masks, foreground, background, and the image in a hierarchy. 
Figure~\ref{fig:overview} gives an overview of our method. 
In a second stage, we generate mask-image pairs to train the Deeplab V3 \cite{chen2017rethinking} segmentation network, thereby enabling unsupervised part segmentation.
Our core network architecture design principle is to build a hierarchy that preserves the translation equivariance of its part representations at each of the following three stages.

\subsection{Level 1: Point Generation and Part Scale}
\label{sec:kp_generator}
\begin{figure}[t]
\begin{center}
   \includegraphics[width=0.98\linewidth]{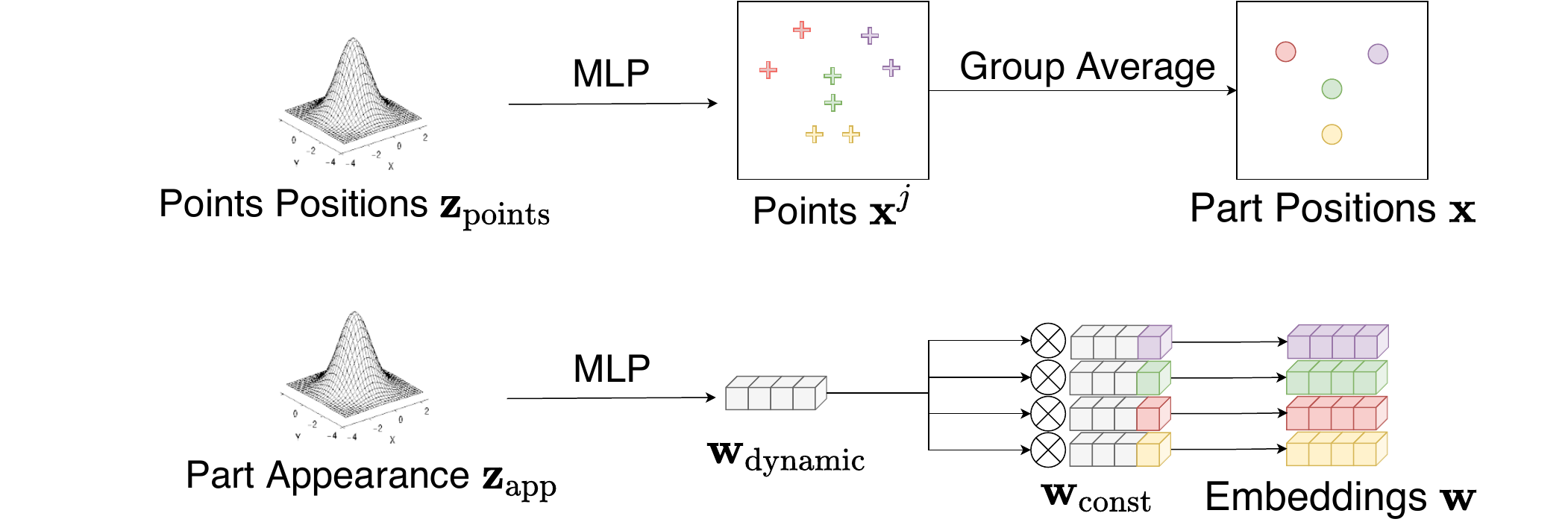}
\end{center}
   \caption{\textbf{Level 1.} The Point Generator uses two Gaussian noise vectors to generate part location, and part appearance embeddings, respectively.}
\label{fig:kp_gen}
\end{figure}

In the first level, we utilize independent noise vectors to generate the locations and appearances of $K$ parts. 
We found training to be most stable by first predicting $n_\text{per} \times K$ points separated into $K$ groups with $n_\text{per}$ points. 
The part location and scale are computed from the mean and standard deviation of the corresponding $n_\text{per}$ points, which regularizes training.
Figure~\ref{fig:kp_gen} gives an overview of the underlying \textbf{Point Generator} module. 
It takes two noise vectors as input $\vz_\text{point},\vz_\text{app}\sim \mathcal N(\mathbf 0^{D_\text{noise}}, \mI^{D_\text{noise}\times D_\text{noise}})$, where $D_\text{noise}$ is the noise dimension. 
We use a 3-layer multi-layer perceptron (MLP) to map $\vz_\text{point}$ to $n_\text{per}\times K$ points $\{\vx_k^{1}, ...,\vx_k^{n_\text{per}}\}_{k=1}^K$. 
Then we calculate the part locations $\{\vx_1,...,\vx_K\}$ and part scales $\{\sigma_1,...,\sigma_K\}$,
\begin{align}
    \vx_k = \frac{1}{n_\text{per}}\sum_{i=1}^{n_\text{per}}\vx_k^i, \quad \sigma_k  &= \frac{\sqrt{\sum_i^{n_\text{per}}\|\vx_k^i-\vx_k\|^2}}{n_\text{per}-1}, \nonumber\\ 
    \text{with }\{\vx_k^{1}, ...,\vx_k^{n_\text{per}}\}_{k=1}^K &= \text{MLP}_\text{point}(\vz_\text{point}),
    \label{eq:gen_kp}
\end{align}
where $k=1,...,K$.

We use another 3-layer MLP to map $\vz_\text{app}$ to a part appearance vector $\vw^\text{dynamic}\in\R^{D_\text{emb}}$. Following \cite{he2021latentkeypointgan}, we define a constant embedding vector $\vw_k^\text{const}\in\R^{D_\text{emb}}$ for each part. We then perform an elementwise multiplication between $\vw^\text{dynamic}$ and $\vw_k^\text{const}$ to obtain the final part embedding $\vw_k\in\R^{D_\text{emb}}$. That is,
\begin{align}
    \vw^\text{dynamic} &= \text{MLP}_\text{app}(\vz_\text{app}) \\
    \vw_k &= \vw^\text{dynamic} \otimes \vw_k^\text{const}
\end{align}
where $\otimes$ is the elementwise product.
It is important that the noise source for appearance and position is independent to prevent the appearance from interfering with the location information. 

\subsection{Level 2: From Points to Masks}
In the second level, the mask generation, we use Gaussian heatmaps to model the local independence and positional encoding \cite{park2019semantic} to generate masks relative to the predicted part location. 
We encode relative instead of absolute position between the points and the image pixels to keep long-distance relations and to prevent leaking the absolute coordinate information, which would violate the translation equivariance. 
To further preserve the translation equivariance, we initialize the positional encoding in a larger grid than the real image range and crop to a fixed margin size after each 2x upsampling \cite{karras2021alias}, which prevents convolutional layers from passing on boundary information (see Section~\ref{sec:lvl3} for additional details). 

\begin{figure}[t]
\begin{center}
   \includegraphics[width=0.98\linewidth]{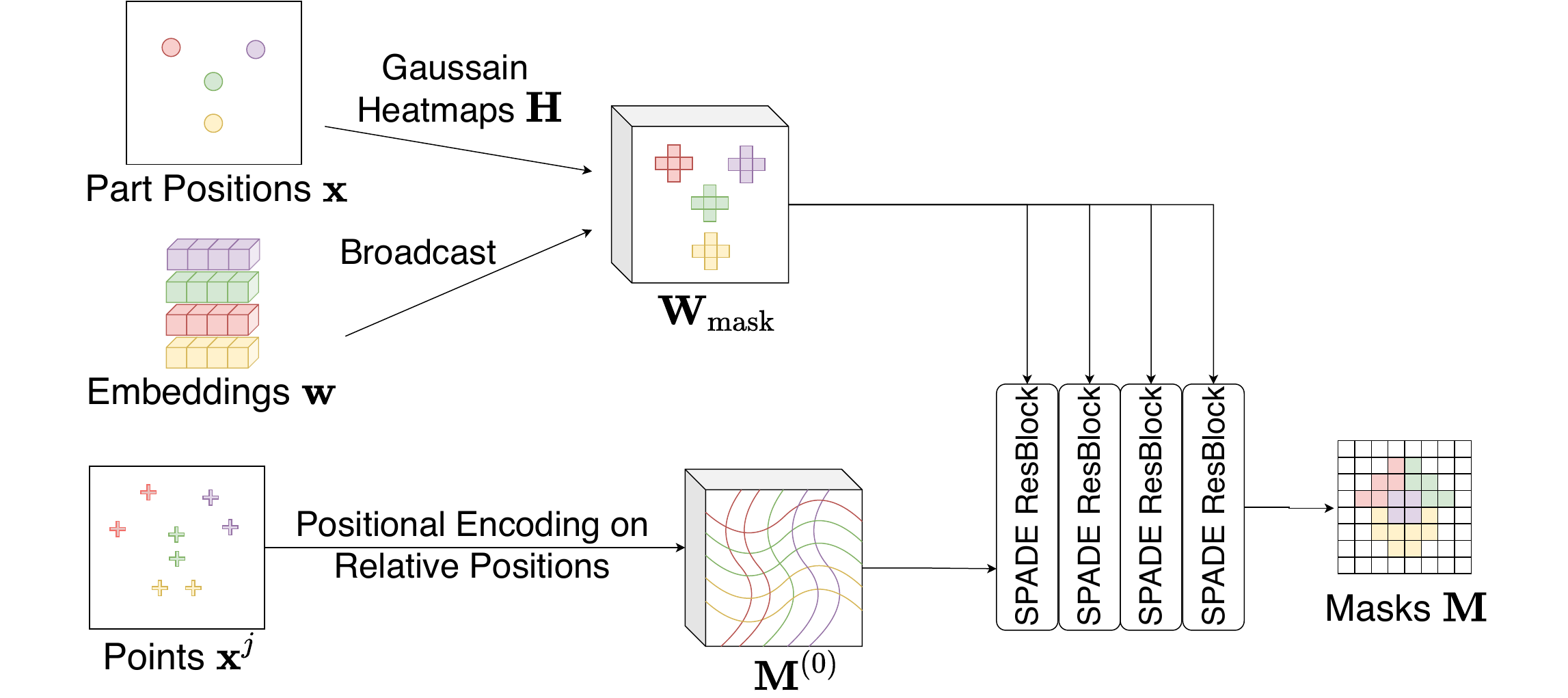}
\end{center}
   \caption{\textbf{Level 2.} The Mask Generator uses points, part locations, part scales, and part appearance embeddings to generate masks.}
\label{fig:mask_gen}
\end{figure}

These operations are implemented with the
\textbf{Mask Generator} \label{sec:mask_generator} illustrated in Figure~\ref{fig:mask_gen}. It takes the $n_\text{per} \times K$ points $\{\vx_k^{1}, ...,\vx_k^{n_\text{per}}\}_{k=1}^K$, part locations $\vx_k$ and part scales $\sigma_K$, and part embeddings $\vw_k$ as input. We generate a Gaussian heatmap for each part using the mean and standard deviation of each part defined in Equation~\ref{eq:gen_kp}. The embedding $\vw_k$ is then multiplied with every pixel of the corresponding heatmap, generating a spatially localized embedding map. We assume the additivity of feature maps (see supplementary for more details). All $K$ part-specific embeddings are summed to form a single feature map $\mW_\text{mask}\in\R^{D_\text{emb}\times H\times W}$. Formally we write,
\begin{equation}
    \begin{split}
        \mH_k(\vp)=\exp&\left(-\|\vp-\vx_k\|_2^2 / \sigma_k^2\right),\\
        \mW_\text{mask}(\vp) &= \sum_{k=1}^K \mH_k(\vp)\vw_k.
    \end{split}
\end{equation}
Note that we use $\sigma_k^2$ instead of $2\sigma_k^2$ to obtain sharper heatmaps, making it easier for the network to generate sharp masks.

The generated embedding map $\mW_\text{mask}$ will subsequently be used to generate masks, together with the mask starting tensor $\mM^{(0)}\in\R^{D_\text{emb}\times H\times W}$. 
To avoid leaking absolute position information, we do not use a constant tensor \cite{karras2019style} or a linearly mapped noise \cite{park2019semantic}. 
Instead, we use low frequency positional encoding \cite{tancik2020fourier} of the difference between the pixel position and the $n_\text{per}\times K$ points. 
That is,
\begin{equation}
    \begin{split}
        \mM^{(0)}(\vp) = [&\sin(\pi\text{FC}([\vp-\vx^1_1, ..., \vp-\vx^{n_\text{per}}_K])),\\
        &\cos(\pi\text{FC}([\vp-\vx^1_1, ..., \vp-\vx^{n_\text{per}}_K]))]
    \end{split}
    \label{eq:positional_encoding}
\end{equation}
where FC stands for a fully connected layer without activation function followed (a linear projection).

With the mask starting tensor $\mM^{(0)}$ and mask embedding map $\mW_\text{mask}$ defined, we generate masks $\mM=[\mM_\text{bg}, \mM_1,...,\mM_K]\in\R^{(K+1) \times H\times W}$ with SPADE ResBlocks~\cite{park2019semantic},
\begin{equation}
    \begin{split}
        \mM^{(i)} = \text{SPADE ResBlock}& (\mM^{(i-1)}, \mW_\text{mask})\\
        \mM = \text{softmax}&(\mM^{(T_\text{mask})})
    \end{split}
\end{equation}
where $i=1,...,T_\text{mask}$, $T_\text{mask}$ is the number of blocks, and an additional channel is reserved for the background.
For more details of SPADE ResBlock, we refer the reader to our supplementary document and the original paper \cite{park2019semantic}.
In theory, the Batch Normalization \cite{ioffe2015batch} may leak absolute position information and break the translation equivariance. 
However, in practice, experiments have shown SPADE has strong local disentanglement \cite{park2019semantic, zhu2020sean, he2021latentkeypointgan}.

\subsection{Level 3: Mask-conditioned Image Generation}
\label{sec:lvl3}
\begin{figure}[t]
\begin{center}
   \includegraphics[width=0.98\linewidth]{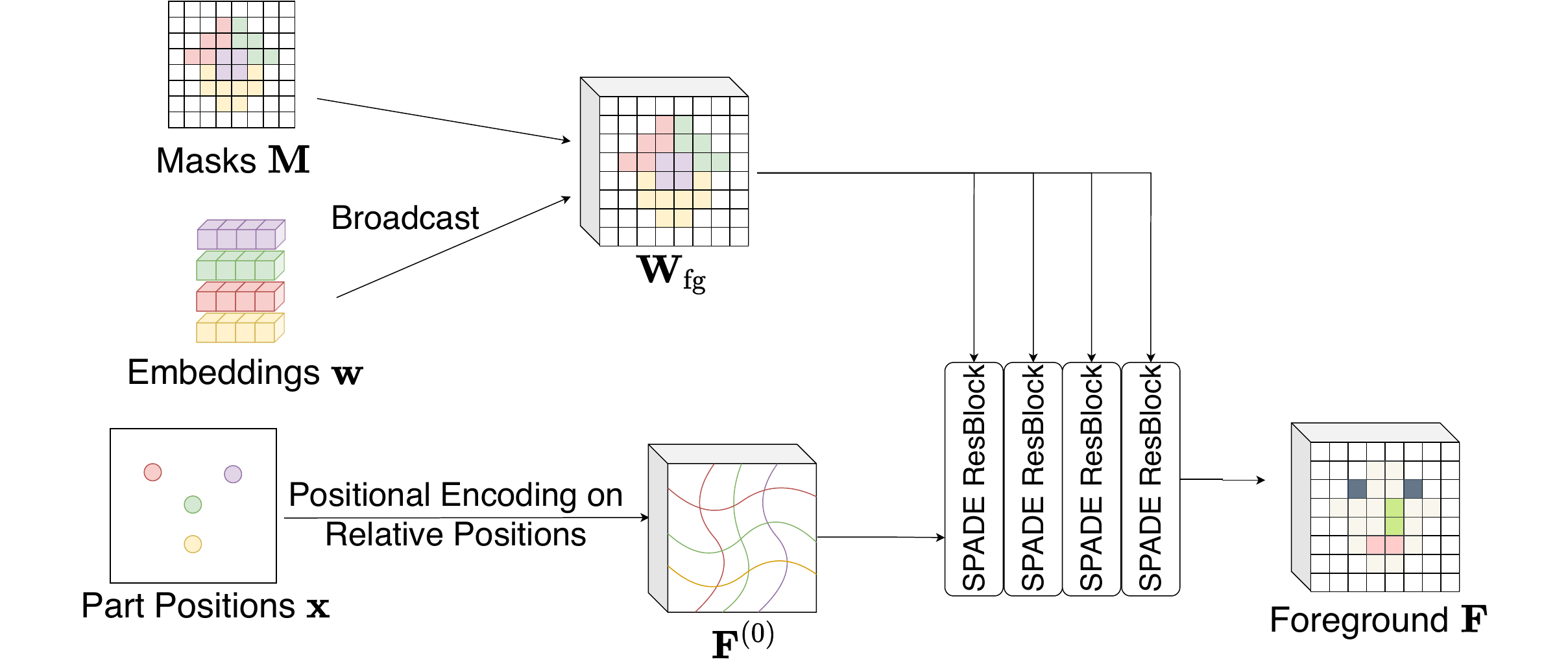}
\end{center}
   \caption{\textbf{Level 3 - Part I.} The foreground Generator uses part locations, part appearance embeddings, and masks.}
\label{fig:fg_gen}
\end{figure}

In the third level, we generate the foreground and the background separately and blend them linearly by reusing the masks from the previous level. 
The \textbf{Foreground Generator} \label{sec:fg_generator} is illustrated in Figure~\ref{fig:fg_gen}. 
It takes the $K+1$ masks $\mM$, $K$ part locations $\vx_k$, and $K$ part appearance embedding $\vw_k$ as input. 
Similar to the procedure that generates masks, we first broadcast the embedding with the corresponding mask to generate the foreground embedding map $\mW_\text{fg}\in\R^{D_\text{emb}\times H\times W}$,
\begin{equation}
    \mW_\text{fg}(\vp) = \sum_{k=1}^K \mM_k(\vp) \vw_k.
\end{equation}

We then use $K$ part locations to generate the foreground starting tensor $\mF^{(0)}$ with low frequency positional encoding similar to Equation~\ref{eq:positional_encoding}.
Finally we use SPADE ResBlocks to generate the Foreground feature map $\mF\in\R^{D_\text{emb}\times H\times W}$,
\begin{equation}
    \begin{split}
        \mF^{(i)} =  \text{SPADE ResBlock} &(\mF^{(i-1)}, \mW_\text{fg})\\
        \mF =  \mF^{(T_\text{fg})},&
    \end{split},
\end{equation}
where $i=1,...,T_\text{fg}$ and $T_\text{fg}$ is the number of SPADE ResBlocks.
Independent of this, the \textbf{Background Generator} \label{sec:bg_generator} takes two noise vectors as input $\vz_\text{bg\_app}\sim \mathcal N(\mathbf 0^{D_\text{noise}}, \mI^{D_\text{noise}\times D_\text{noise}})$, $\vu_\text{bg\_pos}\sim \mathcal U([-1,1]^2)$. We first use a 3-layer MLP to map $\vz_\text{bg\_app}$ to a background appearance vector $\vw_\text{bg}\in\R^{D_\text{emb}}$,
\begin{equation}
    \vw_\text{bg} = \text{MLP}_\text{bg\_app}(\vz_\text{bg\_app}).
\end{equation}
Positional encoding is used on the difference between the background center $\vu_\text{bg\_pos}$ and the pixel position, to generate the background starting tensor $\mB^{(0)}$, similar to Equation~\ref{eq:positional_encoding}.

\begin{figure}[t]
\begin{center}
   \includegraphics[width=0.98\linewidth]{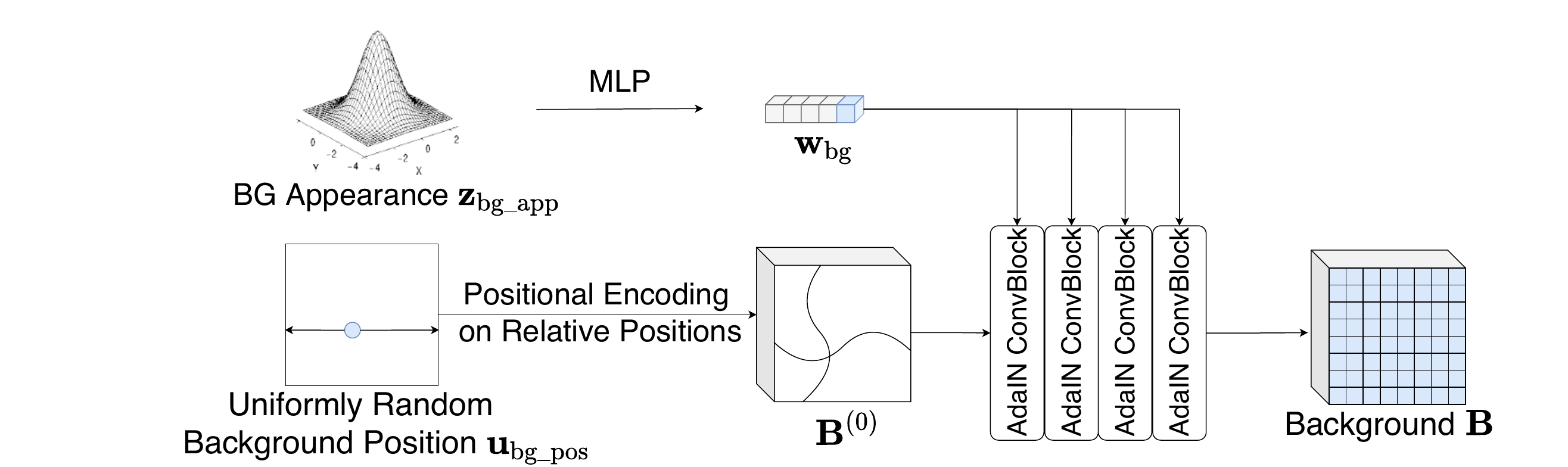}
\end{center}
   \caption{\textbf{Level 4 - Part II.} The Background Generator uses a Gaussian noise vector, and a random position to generate translation-invariant background.}
\label{fig:bg_gen}
\end{figure}

Finally we use AdaIN ConvBlocks \cite{huang2017arbitrary} to generate the background feature map $\mB\in\R^{D_\text{emb}\times H\times W}$,
\begin{equation}
    \begin{split}
        \mB^{(i)} = \text{AdaIN ConvBlock} & (\mB^{(i-1)}, \mW_\text{bg})\\
        \mB = \mB^{(T_\text{bg})}&
    \end{split}
\end{equation}
where $i=1,...,T_\text{bg}$ and $T_\text{bg}$ is the number of AdaIN ConvBlocks.
For more detail on AdaIN ConvBlocks, we refer readers to our supplementary, or the original paper \cite{huang2017arbitrary}.

\noindent\parag{Combining Foreground and Background.} \label{sec:com_fg_bg} Recall that we generate the background mask $\mM_\text{bg}\in\R^{H\times W}$ along with the part masks. This is used to combine our Foreground and Background together. The final image is generated by feeding the feature map into a two-layer CNN. That is,
\begin{equation}
    \mI = \text{Conv}((1-\mM_\text{bg})\otimes F + \mM_\text{bg}\otimes B)
\end{equation}
where $\otimes$ is the pixel-wise product.

\noindent\parag{Upsampling and Cropping.} \label{sec:crop_margin} For simplicity, we used size $H\times W$ for all above feature maps. 
For mask, foreground and background generation, the starting tensor has a 10-pixel-wide margin at each border, same as \cite{karras2021alias}, so that the central feature map will not be interfering with the boundary. For example, instead of generating a $H_0\times W_0$ grid in range $[-1,1]^2$, we generate a $(H_0+20)\times (W_0+20)$ grid in range 
\begin{equation}
        [-1-20/H_0, 1+20/H_0] \times
        [-1-20/W_0, 1+20/W_0] .
\end{equation}
We use this grid to calculate the starting tensors for masks, foreground, and background. After each SPADE ResBlock and each AdaIN ConvBlock, we use 2x upsampling on the feature maps. The margin becomes 20-pixel wide. To bound the otherwise increasing boundary width, we subsequently crop the feature map to keep the 10-pixel margin. The Gaussian heatmaps are calculated on a grid with a 10-pixel-wide-margin, separately for each resolution.

\subsection{Training Objective} \label{sec:loss}

Our hierarchical GAN is trained end-to-end on image collections using the following loss functions.

\noindent\parag{Adversarial Loss.} We denote $\cG$ as the generator and $\cD$ as the discriminator. We use the non-saturating loss \cite{goodfellow2014generative},
\begin{equation}
    \cL_\text{GAN}(\cG)=\mathbb E_{\vz\sim\cN}\log(\exp(-\cD(\cG(\vz)))+1)
\label{eq:gen_loss}
\end{equation}
for the generator, and logistic loss,
\begin{equation}
    \begin{split}
        \cL_\text{GAN}(\cD)=&\mathbb E_{\vz\sim\cN}\log(\exp(\cD(\cG(\vz)))+1)+\\
        &\mathbb E_{\vx\sim p_\text{data}}\log(\exp(-\cD(\vx))+1)
    \end{split}
    \label{eq:dis_loss}
\end{equation}
for the discriminator, with gradient penalty~\cite{mescheder2018training} applied only on real data,
\begin{equation}
    \cL_\text{gp}(\cD)=\mathbb E_{\vx\sim p_\text{data}}\nabla\cD(\vx).
\label{eq:gradient penalty}
\end{equation}

\noindent\parag{Geometric Concentration Loss} Pixels from the same segment are usually connected and concentrated around its center as assumed by \cite{hung2019scops}. We enforce the mask to be in an area around its center with the geometric concentration loss
\begin{equation}
    \cL_\text{con}(\cG)=\sum_{k=1}^K\sum_\vp\frac{\mM_k(\vp)}{\sum_{\vp'} \mM_k(\vp')}\|\vp - \vx_k\|_2^2.
\label{eq:con_loss}
\end{equation}
Note that the background mask is not constrained by $\cL(\cG)_\text{con}$. We found that this loss alone can cause parts to collapse since it encourages a small mask area. To mitigate this problem, we introduce the Area Loss below.

\noindent\parag{Area Loss.} We force the mask area to be larger than the area of the Gaussian heatmap, which is not predefined but predicted from the generated points in level 1. If the part is visible, this loss encourages the visibility of the mask. Otherwise, the area loss encourages a smaller (close to a zero area) Gaussian heatmap.
\begin{equation}
    \cL_\text{area}(\cG)=\sum_{k=1}^K \max\left(0, \sum_\vp\mH_k(\vp)-\sum_\vp\mM_k(\vp)\right).
\label{eq:area_loss}
\end{equation}
We will empirically show that this loss makes the masks more consistent.

The final loss for the discriminator is 
\begin{equation}
    \cL(\cD)=\cL_\text{GAN}(\cD) + \lambda_\text{gp}\cL_\text{gp}(\cD),
\end{equation}
and the final loss for the generator is
\begin{equation}
    \cL(\cG)=\cL_\text{GAN}(\cG) + \lambda_\text{con}\cL_\text{con}(\cG) + \lambda_\text{area}\cL_\text{area}(\cG).
\end{equation}
\section{Experiments} \label{experiment}
Following related work \cite{hung2019scops, liu2021unsupervised, siarohin2021motion}, we analyze the improvement of our method in terms of the positioning of parts and the mask coverage on the established benchmarks. 
We provide results for a wide variety of images containing faces, animals, flowers, and humans.
The supplemental provides additional examples.

\subsection{Datasets and Metrics}

Our evaluation follows the dataset-specific protocols and metrics as established in prior work:

\textbf{CelebA-in-the-wild~\cite{liu2015faceattributes}} shows celebrity faces in unconstrained conditions and is used to estimate the consistency of part positions and the mask center of mass. We follow \cite{hung2019scops} removing images with a face-covering fewer than 30\% of the pixel area from the MAFL train and test sets.
As in \cite{hung2019scops, liu2021unsupervised}, we train a linear regression model without bias from the part centers to the ground truth keypoints. The error metric is the landmark regression error in terms of mean $L_2$ distance normalized by inter-ocular distance.
The splits are 45609 images for GAN training, 5379 with keypoint labels for regression, and 283 for testing. 

\textbf{CelebA-aligned} \cite{liu2015faceattributes} contains 200k faces, each centered such that eyes align. Following \cite{thewlis2017unsupervised}, we use three subsets: CelebA training set without MAFL (160k images), MAFL training set (19k images), MAFL test set (1k images). The error metric is the same as on CelebA-in-the-wild.

\textbf{CUB-2011} \cite{WahCUB_200_2011} consists of 11,788 images of birds. We follow \cite{chen2019unsupervised} to use 10,000 images for training, 1,000 for testing, and the remaining 778 for validation. We use this dataset to analyze segmentation coverage accuracy. We aggregate part segments to form the foreground mask and calculate the Intersection Over Union (IoU) between the predicted foreground masks and the ground truth foreground masks \cite{hung2019scops, liu2021unsupervised}. To this end, we calculate the foreground mask as the sum of our part masks. 

\textbf{Flowers} \cite{Nilsback08} consists of 8,189 images of flowers. The ground truth masks are obtained by an automated method built specifically for flowers \cite{Nilsback08}. We follow \cite{chen2019unsupervised} to use 6,149 images for training, 1,020 for testing, and 1020 for validation. The metric is IoU of the foreground mask.

\textbf{Taichi} \cite{liu2015faceattributes} contains 3049 training videos and 285 test videos of people performing Tai-Chi. We train our GAN on the training images (\emph{not} videos) using 5000 images for fitting the regression model and 300 other images for testing. For a fair comparison, we use the same 5300 images as \cite{siarohin2021motion}. The metric, mean average error (MAE), is calculated as the sum of the $L_2$ distances between 18 regressed keypoints and their ground truth. We also calculate IoU based on the provided foreground masks.

\subsection{Baselines} 
We compare to the following unsupervised methods, most requiring stronger assumptions:
\textbf{DEF} \cite{collins2018deep} uses test images at training time, which has a high computational cost. 
\textbf{SCOPS} \cite{hung2019scops} uses saliency maps.
\textbf{Liu et al.} \cite{liu2021unsupervised} requires a strong prior on the mask center being close to the image center.
\textbf{Siarohin} et al. \cite{siarohin2021motion} trained on videos exploiting temporal information.


%

\subsection{Qualitative analysis}
\begin{figure}[t]
\centering
  \resizebox{0.98\linewidth}{!}{%
\begin{tabular}{ccccccc}
\includegraphics[]{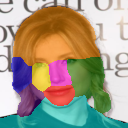}&
\includegraphics[]{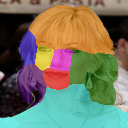}&
\includegraphics[]{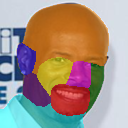}&
\includegraphics[]{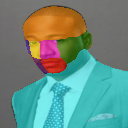}&
\includegraphics[]{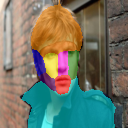}&\\
\includegraphics[]{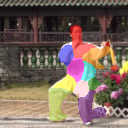}&
\includegraphics[]{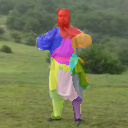}&
\includegraphics[]{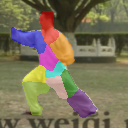}&
\includegraphics[]{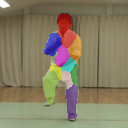}&
\includegraphics[]{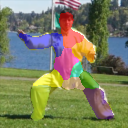}&\\
\includegraphics[]{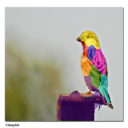}&
\includegraphics[]{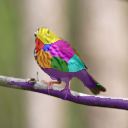}&
\includegraphics[]{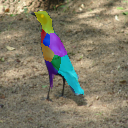}&
\includegraphics[]{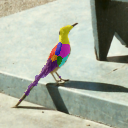}&
\includegraphics[]{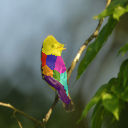}&\\
\includegraphics[]{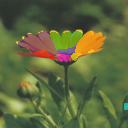}&
\includegraphics[]{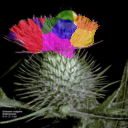}&
\includegraphics[]{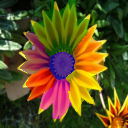}&
\includegraphics[]{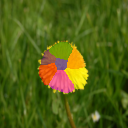}&
\includegraphics[]{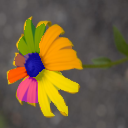}&
\end{tabular}}%
\caption{\textbf{Detection Examples} showing faithful part segmentation on the tested datasets, with varying object size and complexity.}
\label{fig:quality_result}
\end{figure}

\begin{figure}[t]
  \resizebox{0.98\linewidth}{!}{%
\begin{tabular}{cccccccc}
\includegraphics[width=0.5\linewidth]{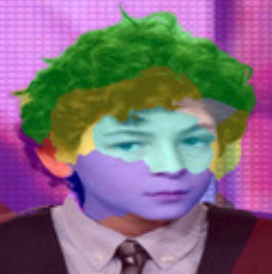}&
\includegraphics[width=0.5\linewidth]{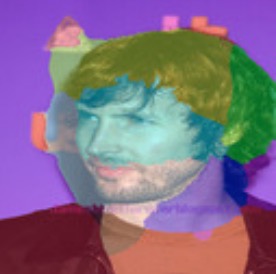}&
\includegraphics[width=0.5\linewidth]{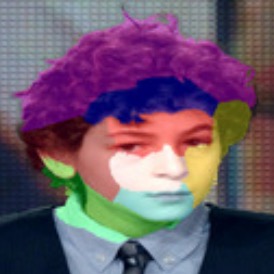}&
\includegraphics[width=0.5\linewidth]{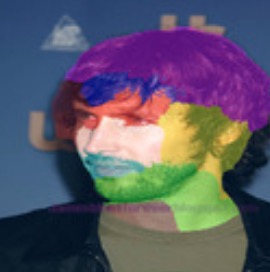}&
\includegraphics[width=0.5\linewidth]{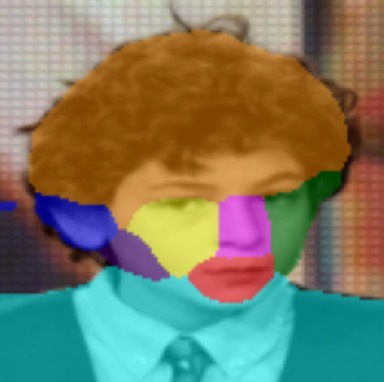}&
\includegraphics[width=0.5\linewidth]{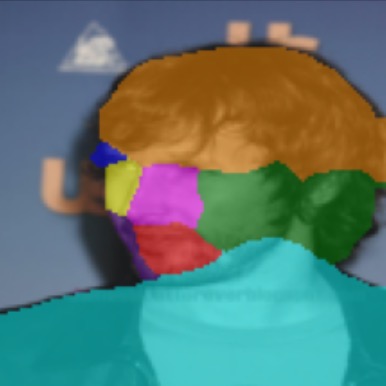}&\\
\includegraphics[width=0.5\linewidth]{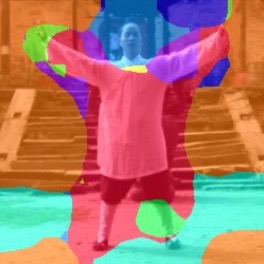}&
\includegraphics[width=0.5\linewidth]{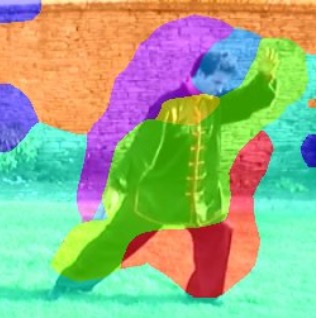}&
\includegraphics[width=0.5\linewidth]{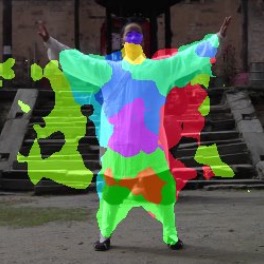}&
\includegraphics[width=0.5\linewidth]{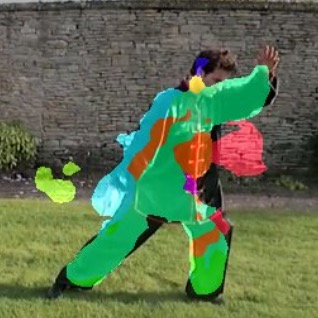}&
\includegraphics[width=0.5\linewidth]{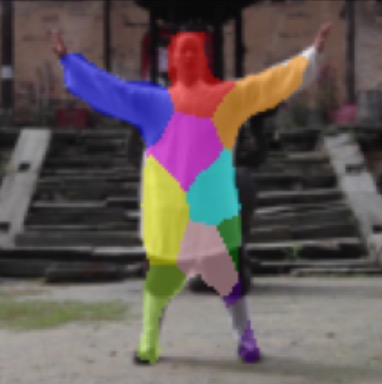}&
\includegraphics[width=0.5\linewidth]{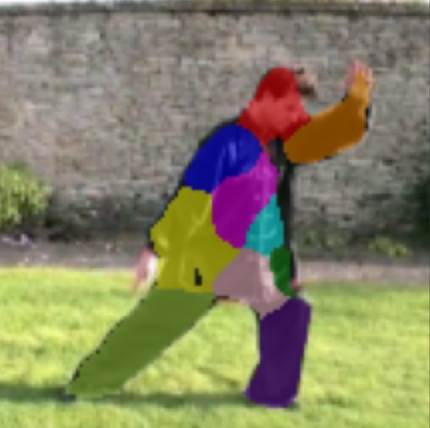}&\\
\rule{0pt}{1pt}\\ 
\multicolumn{2}{c}{{\Huge DEF \cite{collins2018deep}}}  & \multicolumn{2}{c}{{\Huge SCOPS \cite{hung2019scops}}} & \multicolumn{2}{c}{{\Huge\textbf{Ours}}}\\
\end{tabular}}

\caption{\textbf{Mask quality comparison.} We qualitatively compare the mask on wild CelebA (first row) and Taichi (second row). 
Our masks have better quality than other methods due to our hierarchical generator.
}
\label{fig:quality_comparison}
\end{figure}

We show detection examples in Figure~\ref{fig:quality_result}, and qualitatively compare our predicted masks with the baselines in Figure~\ref{fig:quality_comparison}. Our mask has better coverage with the foreground object, less fragmented parts, and better consistency, with the same part masks mapping to the same body parts across different images.
For example, the shoulder exists in almost all images in wild CelebA \cite{liu2015faceattributes}, but SCOPS \cite{hung2019scops} cannot consistently discover this apparent part, likely because their employed saliency map focuses on the face. 
For more qualitative comparisons, refer to the supplementary.

\subsection{Part Center Consistency} \label{sec:part_center_consitency}

\begin{table}[t]
\centering
\resizebox{0.98\linewidth}{!}{%
\begin{tabular}{|l|l|c|c|c|}
\hline
Method & Type & Aligned (K=10) & Wild (K=4) & Wild (K=8)  \\ \hline
Thewlis et al. \cite{thewlis2017unsupervised} & Landmark & 7.95\% & - & 31.30\% $\star$ \\ 
Zhang et al. \cite{zhang2018unsupervised} & Landmark & 3.46\% & - & 40.82\% $\star$\\ 
LatentKeypointGAN \cite{he2021latentkeypointgan} & Landmark & 5.85\% & 25.81\% & 21.90\% \\ 
Lorenz et al. \cite{lorenz2019unsupervised} & Landmark & 3.24\% & 15.49\% $\dagger$ & 11.41\% $\dagger$\\
IMM \cite{jakab2018unsupervised} & Landmark &  \textbf{3.19}\% & 19.42\% $\dagger$ & 8.74\% $\dagger$\\ \hline
DEF \cite{collins2018deep} by \cite{hung2019scops} & Part & - & - & 31.30\% $\star$ \\
SCOPS \cite{hung2019scops} (w/o saliency)  & Part & - & 46.62\% & 22.11\% \\
SCOPS \cite{hung2019scops} (w/ saliency)  & Part & - & 21.76\% & 15.01\% \\
Liu et al. \cite{liu2021unsupervised}  & Part & - & 15.39\% & 12.26\% \\
Huang et al. \cite{huang2020interpretable} (w/ detailed label)  & Part & - & - & 8.40\% \\
\textbf{Ours} & Part & 3.98\% & \textbf{12.26}\% & \textbf{6.18}\% \\ \hline
\end{tabular}}
\caption{\textbf{Landmark detection on CelebA}. The metric is the landmark regression (without bias) error in terms of mean $L_2$ distance normalized by inter-ocular distance (lower is better). Although the pair-based methods work better on aligned CelebA, they do not generalize well. Our generation-based methods are more robust on Wild CelebA. The sign $\star$ means reported by \cite{hung2019scops} and $\dagger$ by \cite{liu2021unsupervised}.} 
\label{tab:keypoint_detection}
\end{table}

Table~\ref{tab:keypoint_detection} shows the results for keypoint detection.
Our method outperforms other part segmentation methods on the part consistency metric on the challenging CelebA-in-the-wild. The keypoint-based methods that perform better on aligned face images do not generalize to real-world examples. The results on Taichi in Table~\ref{tab:taichi} confirm that our model is comparable in consistency (MAE metric) eventhough others are trained with additional information. In addition, we evaluate our model on CUB using the protocol of \cite{lorenz2019unsupervised}. Our estimation error 3.23\% ($L_2$ error normalized by edge length of the image) is lower (better) than Zhang et al. \cite{zhang2018unsupervised} (5.36\%) and Lorenz et al. \cite{lorenz2019unsupervised} (3.91\%).

\subsection{Mask Coverage} \label{sec:mask_coverage}
Since the datasets used in \cite{hung2019scops, liu2021unsupervised} are too small for training a GAN (see limitations) we only use the sufficiently large CUB-2011 \cite{liu2015faceattributes} and Flowers \cite{Nilsback08} dataset.
To nevertheless compare with SCOPS \cite{hung2019scops} we retrain their model.
We use the ground truth masks and the required saliency masks on our train/val/test split with their default parameters. Furthermore, we compare our results with the unsupervised foreground-background segmentation methods that already report on these datasets. 
Note that these methods focus on foreground-background segmentation instead of part segmentation. Their approaches are tailored for this task while it is rather a byproduct for us.

\begin{table}[t]
\centering
\resizebox{0.98\linewidth}{!}{%
\begin{tabular}{|l|l|c|c|}
\hline
Method & Type & CUB & Flowers \\ \hline
GrabCut \cite{rother2004grabcut} & Foreground Segmentation & 0.360 & 0.692 \\ 
PerturbGAN \cite{bielski2019emergence} & Foreground Segmentation & 0.380 & - \\ 
ReDO \cite{chen2019unsupervised} & Foreground Segmentation & 0.426 & 0.764 \\ 
Savarese et al. \cite{savarese2021information} & Foreground Segmentation & \textbf{0.551} & \textbf{0.789} \\ \hline
SCOPS \cite{hung2019scops} w/ GT BG by us& Part Segmentation & 0.329 $\dagger$ & 0.544 $\dagger$ \\
\textbf{Ours} & Part Segmentation & \textbf{0.629} & \textbf{0.739}\\ \hline
\end{tabular}}
\caption{\textbf{Foreground-Background Segmentation}. The metric on CUB and Flowers is the IoU of foreground (higher is better). We use $K=8$ for both ours and SCOPS \cite{hung2019scops}. The sign $\dagger$ means being trained by us using their official implementation.}
\label{tab:fg_iou}
\end{table}

As shown in Table~\ref{tab:fg_iou}, our model outperforms the unsupervised part segmentation methods with better mask coverage, and is comparable with the dedicated foreground segmentation methods. On the most challenging Taichi dataset, we outperform all methods in Table~\ref{tab:taichi} in terms of mask coverage (foreground IOU) even though they use additional information, such as saliency maps and videos.


\begin{table}[t]
\centering
\resizebox{0.95\linewidth}{!}{%
\begin{tabular}{|l|l|c|c|}
\hline
Method & Type & MAE $\downarrow$ & IoU $\uparrow$ \\ \hline
DFF \cite{collins2018deep}& no train/test split  & 494.48 $\dagger$ & -\\
SCOPS \cite{hung2019scops}& images with saliency maps & 411.38 $\dagger$ &  0.5485 $\dagger$\\
Siarohin et al. \cite{siarohin2021motion} & trained on videos & \textbf{389.78} &  0.7686\\
\textbf{Ours} & trained on single images & 417.17 & \textbf{0.8538}\\ \hline
\end{tabular}}
\caption{\textbf{Part Segmentation on Taichi}. The sign $\dagger$ means being reported by \cite{siarohin2021motion}. All the results are shown in the case of $K=10$.}
\label{tab:taichi}
\end{table}

\subsection{Ablation Tests}
\begin{figure}[t]
\begin{center}
   \includegraphics[width=0.85\linewidth]{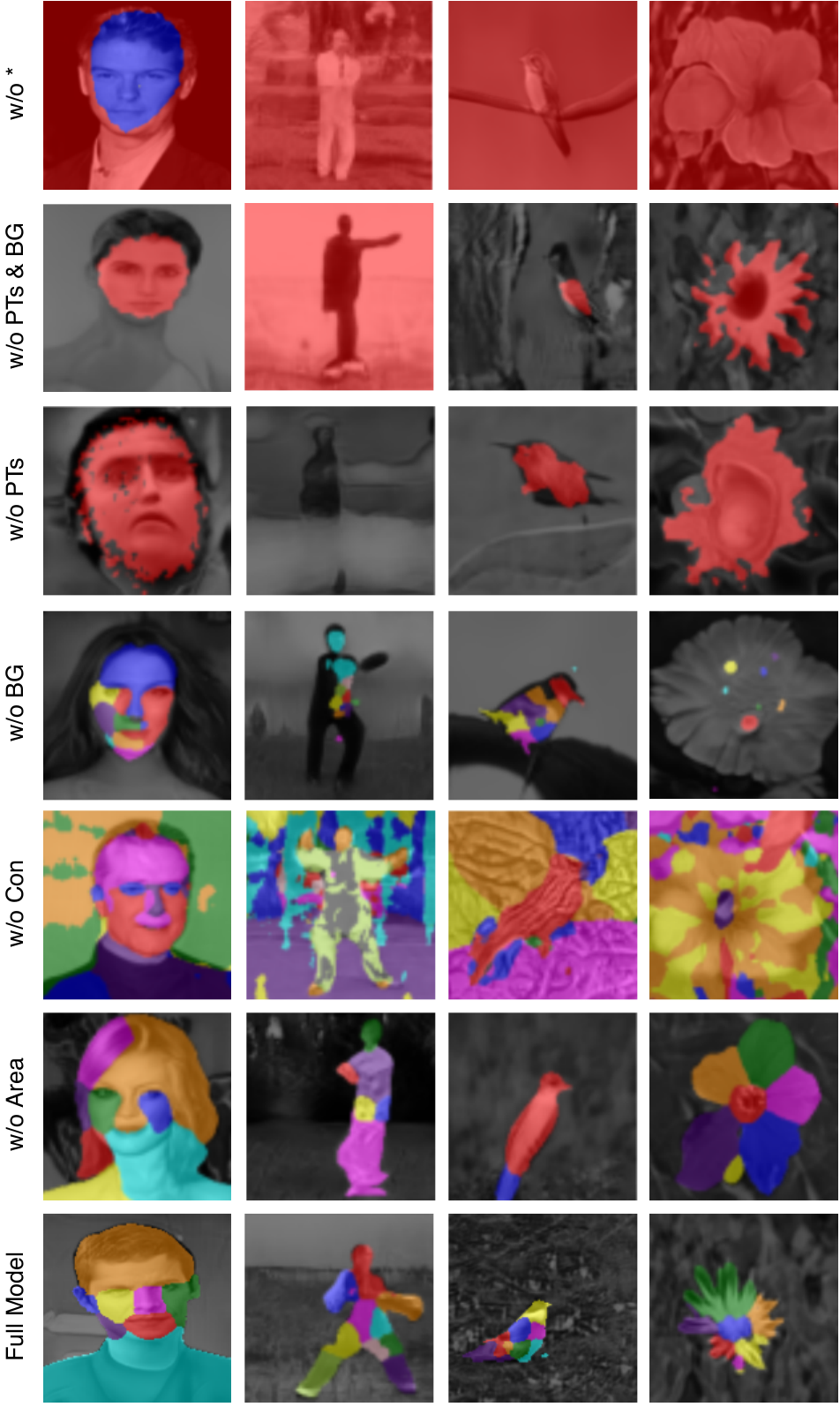}
\end{center}
\vspace{-10pt}
   \caption{\textbf{Ablation Tests.} From top to bottom: models disabling 1) points, 2) separated background, 3) points and separated background, 4) geometric concentration loss 5) area loss, and 6) nothing (full model). For visualization purposes we show the background in grey-scale. All contributions improve across datasets.} 
\label{fig:ablation}
\end{figure}

Figure~\ref{fig:ablation} shows a qualitative comparison when removing different parts from our full model. 
Instead of showing the segmentation on real images, we show the generated masks from our GAN. These masks explain our model more directly.
For ablation tests on the number of parts, we refer readers to the supplementary document.

\textbf{Disabling Points.} We remove the point generation from the network and in turn also the Gaussian heatmaps and SPADE blocks. Instead, we use AdaIN \cite{huang2017arbitrary}, which is a spatially invariant version of SPADE. We use the mean of the part embeddings as the embedding for the feature map of AdaIN and replace the starting tensor with a learned constant tensor as in StyleGAN \cite{karras2019style}. 

\textbf{Disabling Separated Background Generation.} We integrate the background generator into the foreground generator and treat the background as one of the parts. 

\textbf{Disabling Losses.} We remove losses and show examples from each dataset in Figure~\ref{fig:ablation}. Without the Geometric Concentration Loss, masks become fragmented. Without the Area Loss some parts vanish.
Surprisingly, even without points and separate background generation, the model can discover some commonly shared parts in the images except for human bodies, as shown in the third row in Figure~\ref{fig:ablation}. We hypothesize it is due to our spatially variant network design with translation equivariance encouraged at all levels.
Overall, 
it is evident that our hierarchical architecture is crucial for mask generation. 
\section{Limitations and Future Work}
\begin{figure}[t]
\centering
  \resizebox{0.98\linewidth}{!}{%
\begin{tabular}{ccccc}
\includegraphics[width=0.5\linewidth]{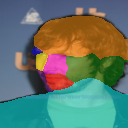}&
\includegraphics[width=0.5\linewidth]{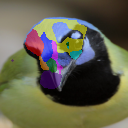}&
\includegraphics[width=0.5\linewidth]{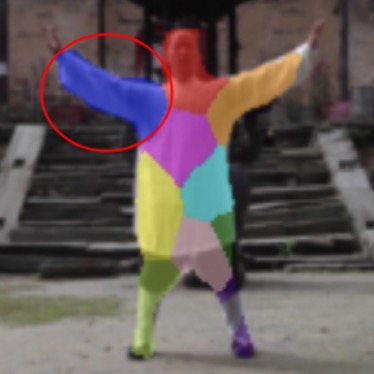}&
\includegraphics[width=0.5\linewidth]{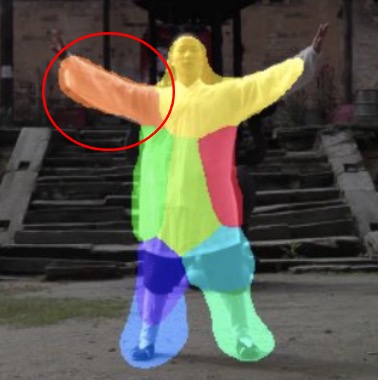}&
\includegraphics[width=0.5\linewidth]{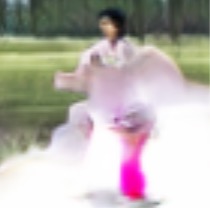}\\
\includegraphics[width=0.5\linewidth]{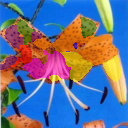}&
\includegraphics[width=0.5\linewidth]{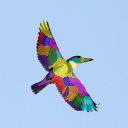}&
\includegraphics[width=0.5\linewidth]{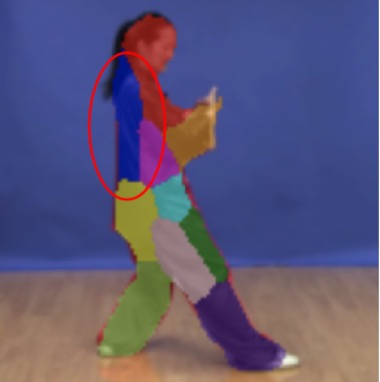}&
\includegraphics[width=0.5\linewidth]{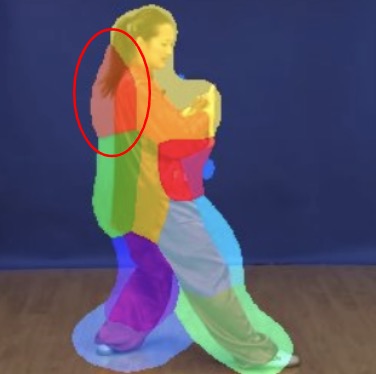}&
\includegraphics[width=0.5\linewidth]{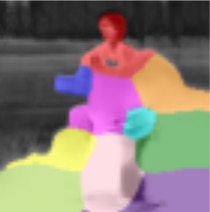}\\
\multicolumn{2}{c}{ } & \huge Ours  & \huge Siarohin et al. \cite{siarohin2021motion}\\
\end{tabular}}%
\caption{\textbf{Limitations}. The method fails for poses underrepresented in the training set (1st and 2nd col) and under occlusions (left eye of the face is marked as blue, which is usually on the ear), which we have in common with other methods such as \cite{siarohin2021motion} (4th col). In some rare cases the generator fails to generate good body shape on Taichi (5th col).}
\label{fig:limitation}
\end{figure}

Our GAN training requires a larger dataset ($>$5000 images), which does not apply to some existing benchmarks but is acceptable for domains with large unlabelled image collections. 
Figure~\ref{fig:limitation} shows representative failure cases that are typical also for related work. In some rare cases and only for the Taichi dataset, the generator fails to generate good shape. Nevertheless, the detection remains accurate as there is no such unusual shape in the real data. When a part is occluded in the image the associated mask will still cover a region nearby, such as the left arm moving to the back instead of being occluded. Sometimes, front/back for bird and left/right for humans are flipped.
The comparison to  \cite{siarohin2021motion} shows that this is common for 2D methods and demands for extensions to 3D and occlusion handling.

\section{Conclusion}

We presented a GAN-based approach for learning part segmentation from unlabelled image collections. Crucial is our hierarchical generator design that synthesizes images in a coarse-to-fine manner, with independence and invariance built in. It forms a viable alternative to existing autoencoder techniques and opens up a path for learning part-based 3D models from 2D images in the future.

\section*{Acknowledgement}
This work was supported by the UBC Advanced Research Computing (ARC) GPU cluster, the Compute Canada GPU servers, and a Huawei-UBC Joint Lab project.

{\small
\bibliographystyle{ieee_fullname}
\bibliography{egbib}
}

\clearpage
\appendix

\section{Additional Results}

\begin{figure*}[t]
\begin{center}
   \includegraphics[width=0.96\linewidth]{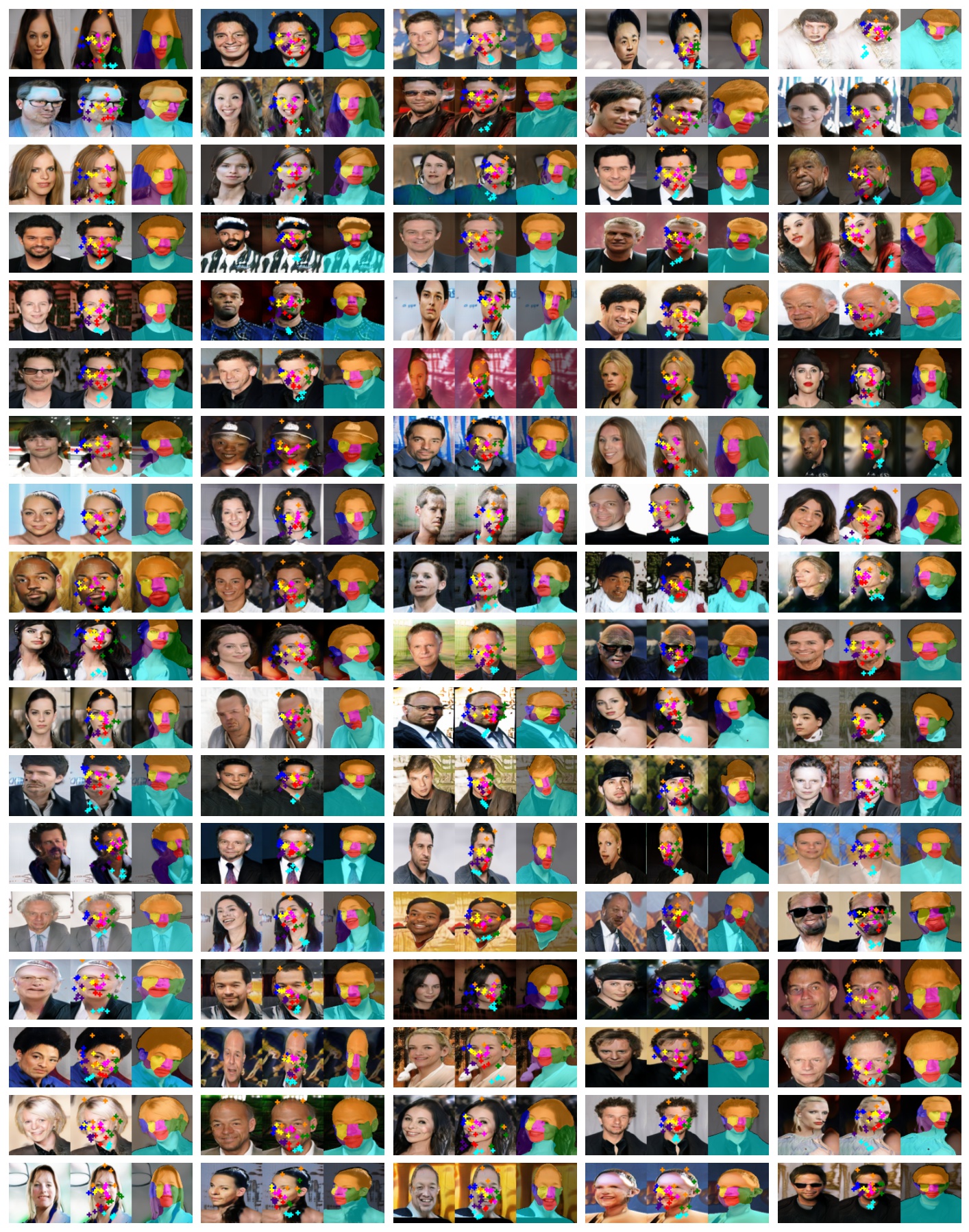}
\end{center}
   \caption{\textbf{90 generated samples from CelebA-in-the-wild,} with the generated image-points-masks pairs overlaid.}
   \label{fig:gen_celeba}
\end{figure*}

\begin{figure*}[t]
\begin{center}
   \includegraphics[width=0.96\linewidth]{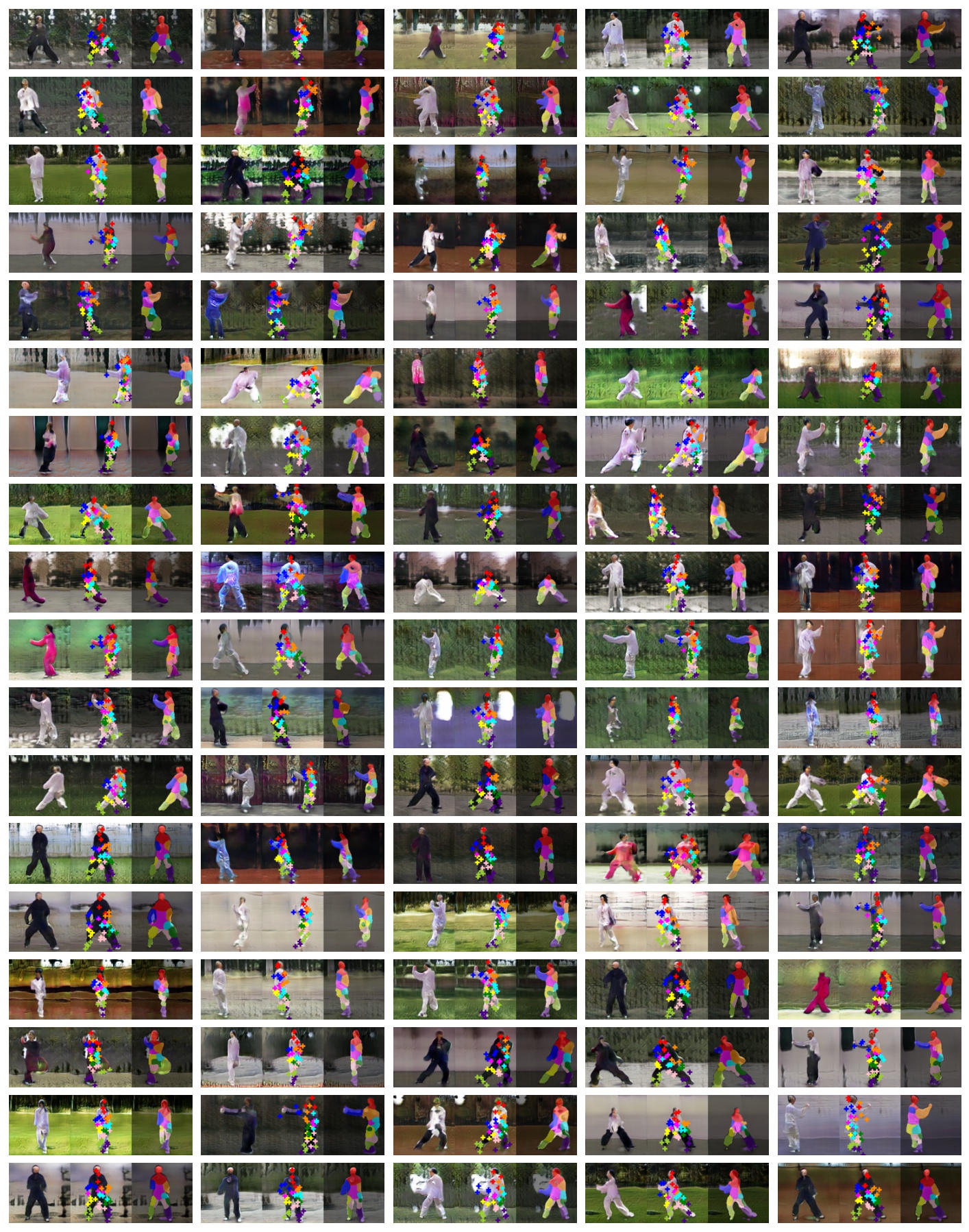}
\end{center}
   \caption{\textbf{90 generated samples from Taichi,} with image-points-masks pairs overlaid.}
   \label{fig:gen_taichi}
\end{figure*}

\begin{figure*}[t]
\begin{center}
   \includegraphics[width=0.96\linewidth]{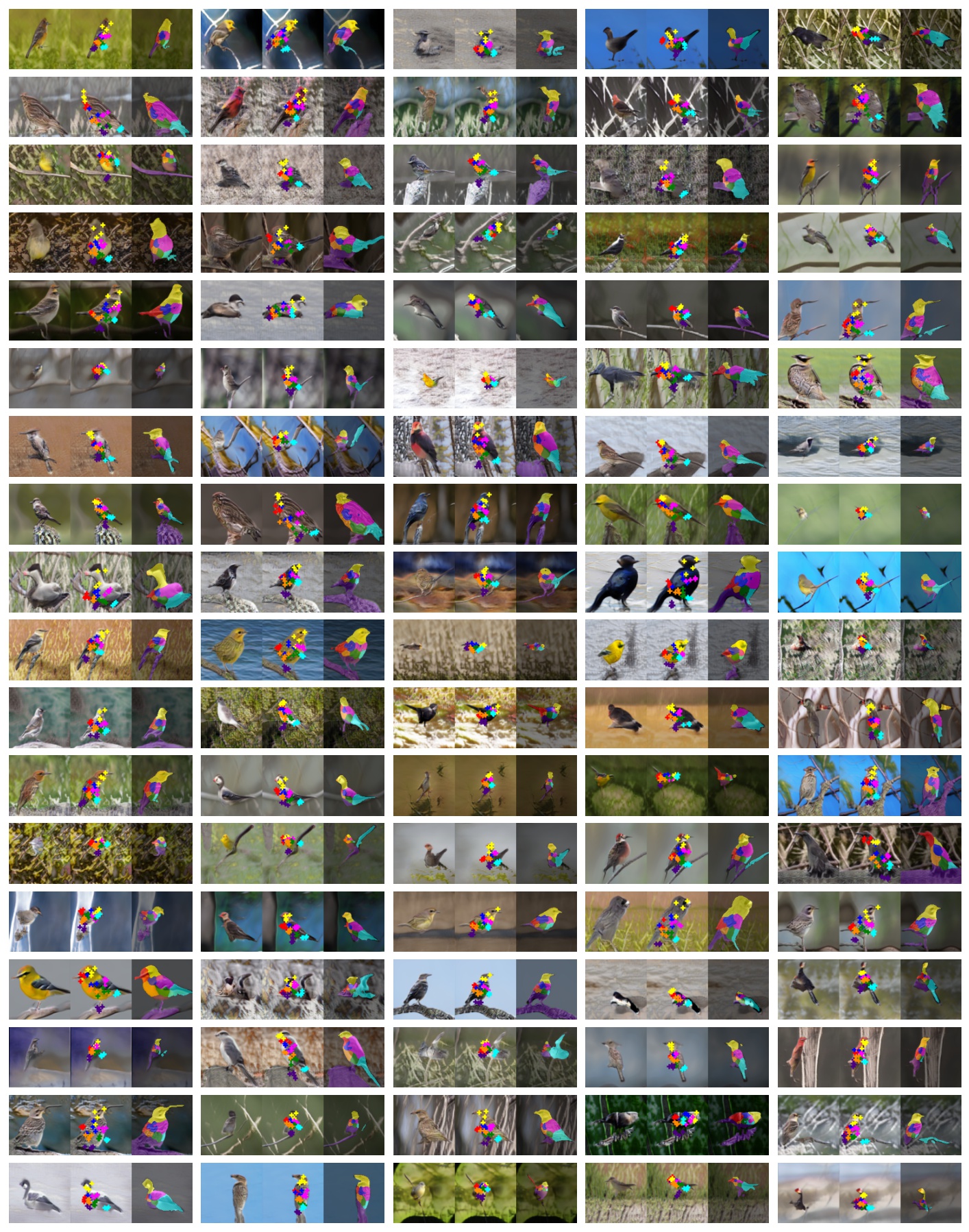}
\end{center}
   \caption{\textbf{90 generated samples from CUB,} with image-points-masks pairs overlaid.}
   \label{fig:gen_cub}
\end{figure*}

\begin{figure*}[t]
\begin{center}
   \includegraphics[width=0.96\linewidth]{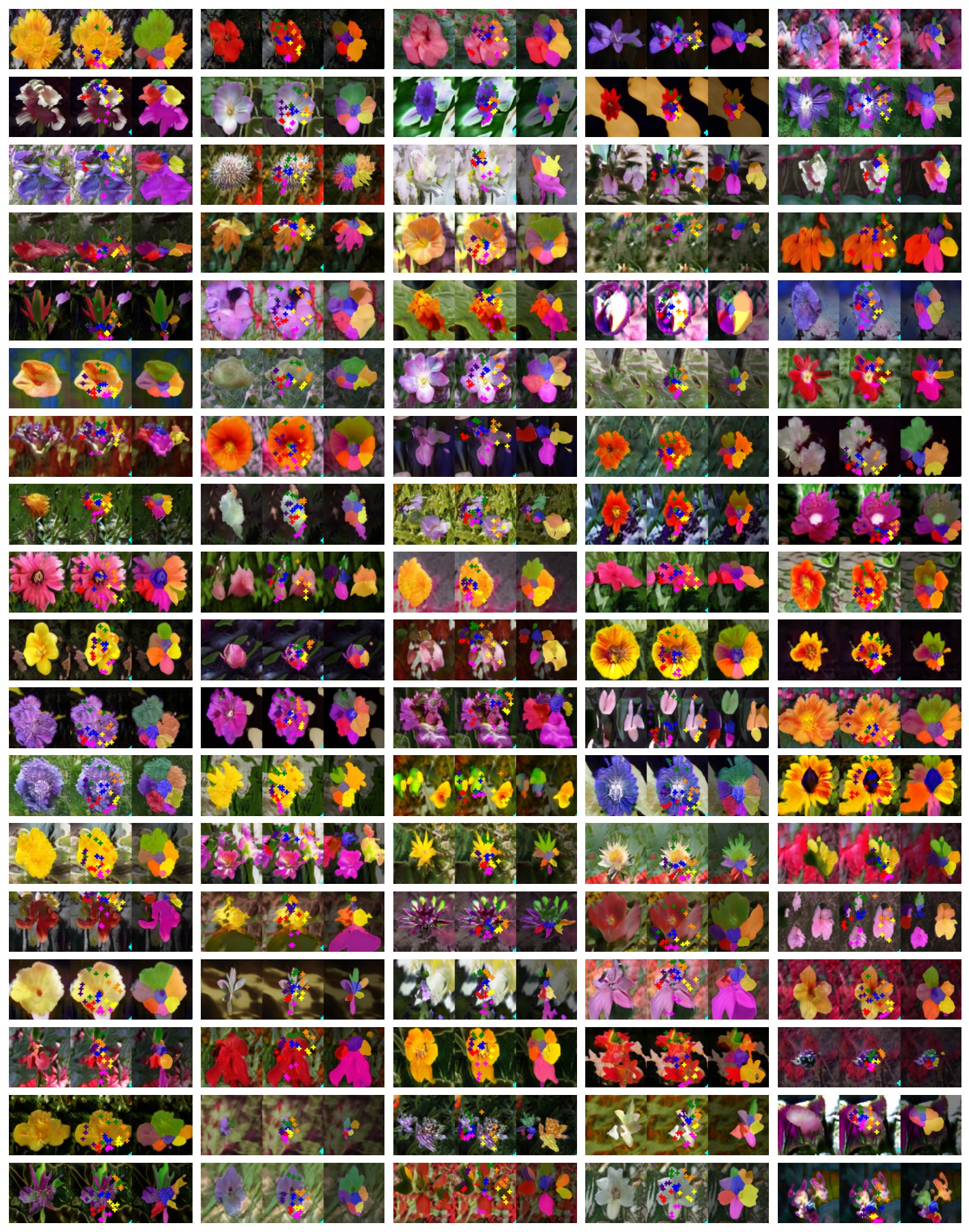}
\end{center}
   \caption{\textbf{90 generated samples from Flowers,} with image-points-masks pairs overlaid.}
   \label{fig:gen_flower}
\end{figure*}

\begin{figure*}[t]
\begin{center}
   \includegraphics[width=0.98\linewidth]{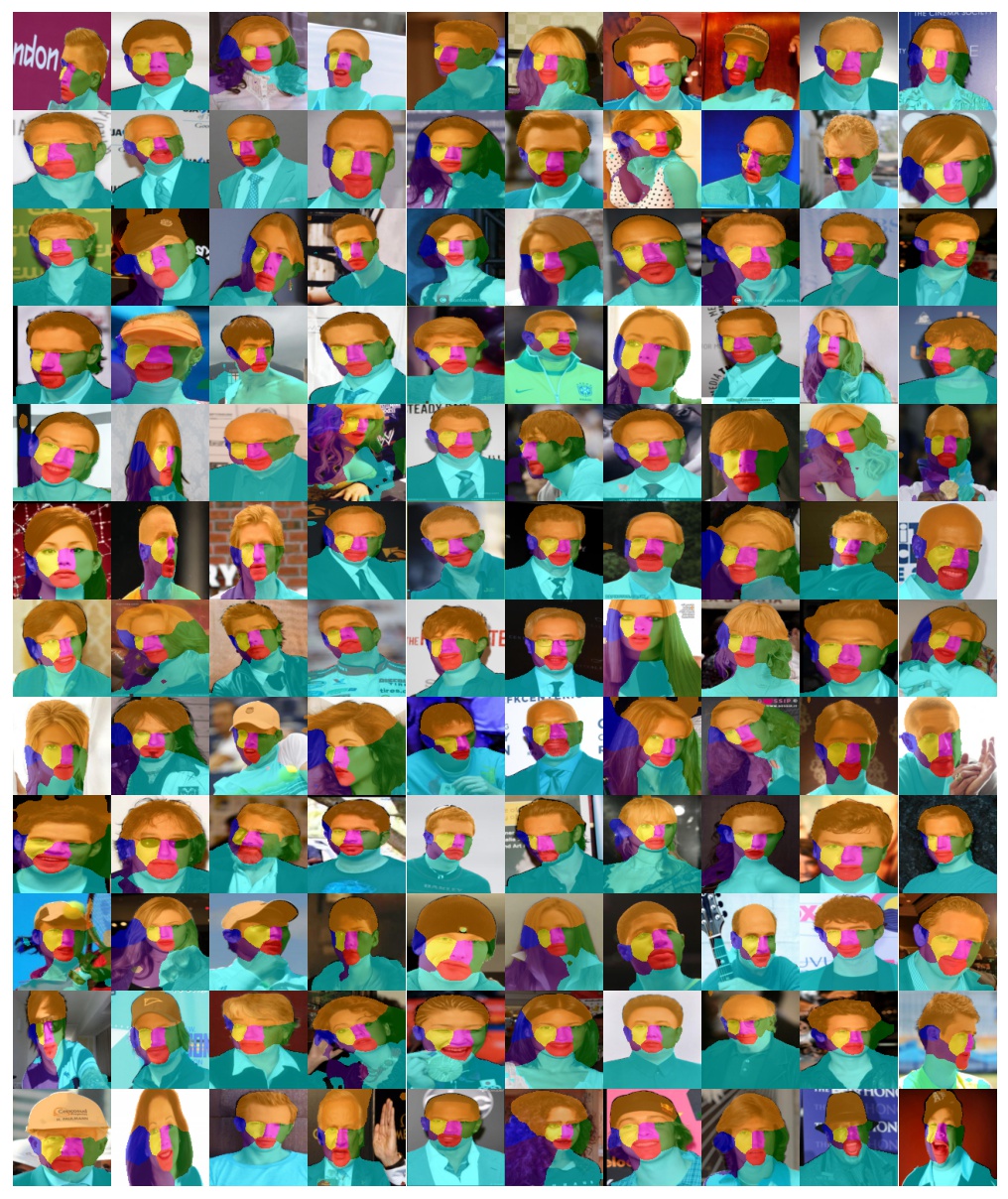}
\end{center}
   \caption{\textbf{120 samples from CelebA-in-the-wild,} with detected masks overlaid.}
   \label{fig:det_celeba}
\end{figure*}

\begin{figure*}[t]
\begin{center}
   \includegraphics[width=0.98\linewidth]{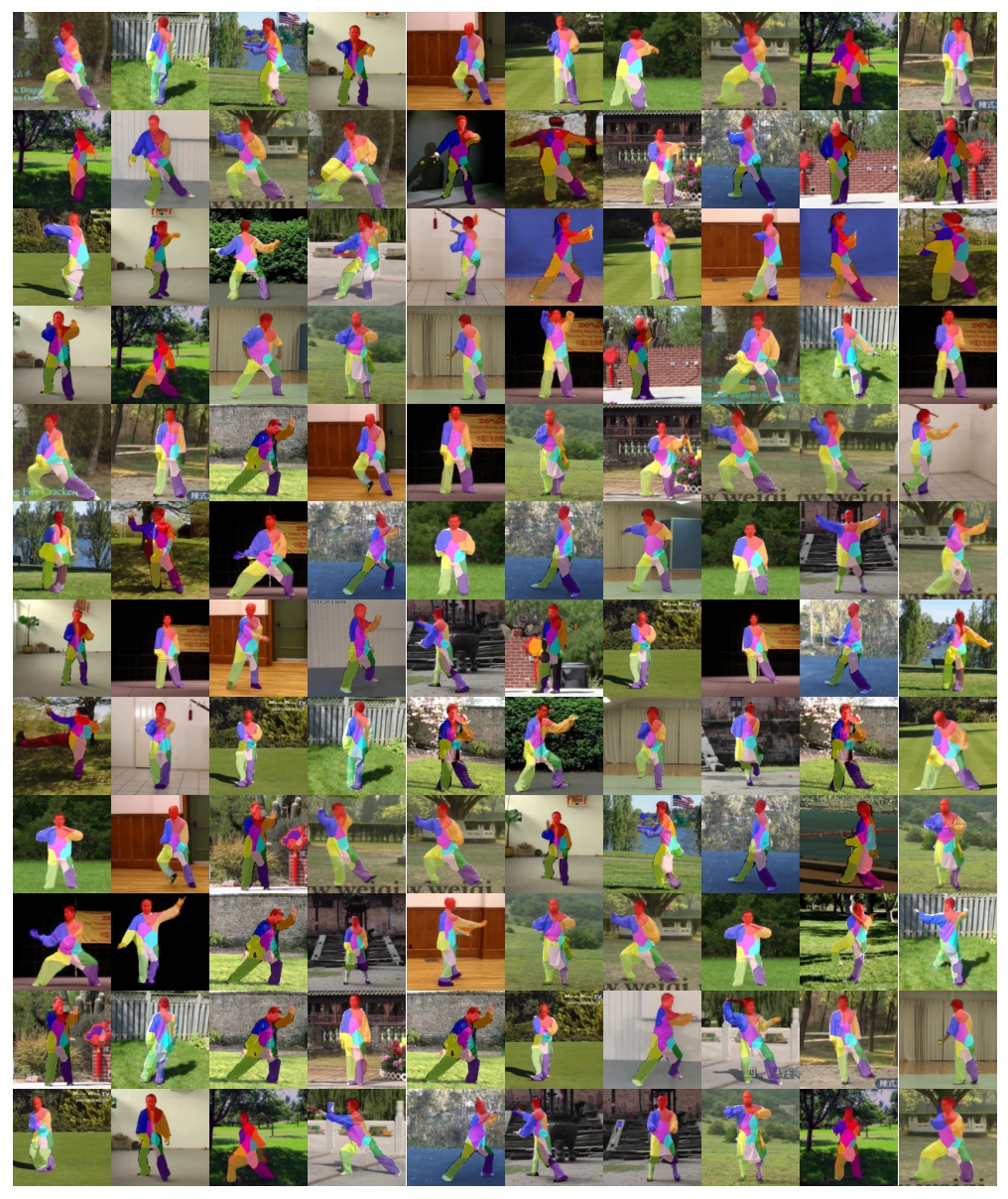}
\end{center}
   \caption{\textbf{120 samples from Taichi,} with detected masks overlaid.}
   \label{fig:det_taichi}
\end{figure*}

\begin{figure*}[t]
\begin{center}
   \includegraphics[width=0.98\linewidth]{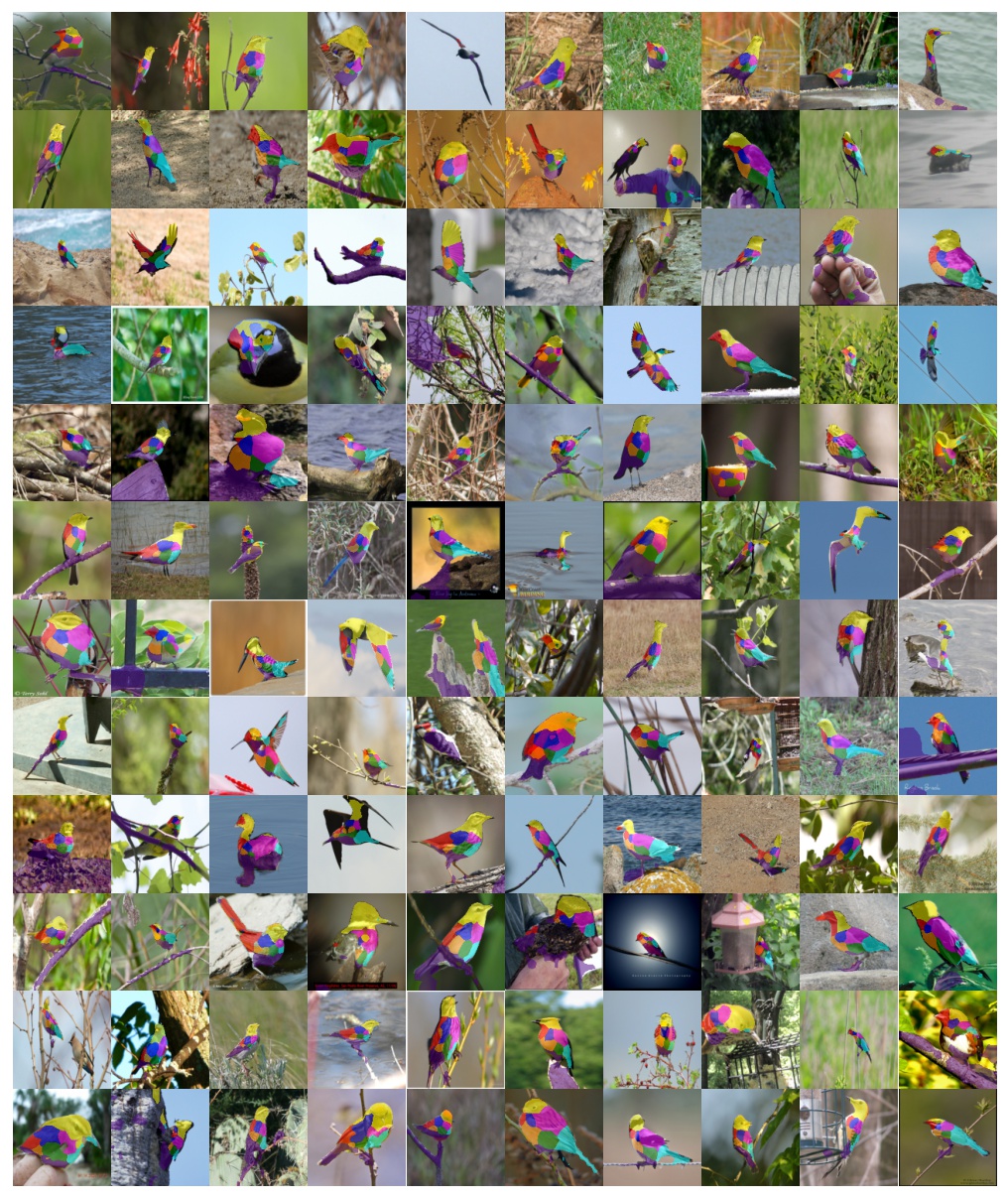}
\end{center}
   \caption{\textbf{120 samples from CUB,} with detected masks overlaid.}
   \label{fig:det_cub}
\end{figure*}

\begin{figure*}[t]
\begin{center}
   \includegraphics[width=0.98\linewidth]{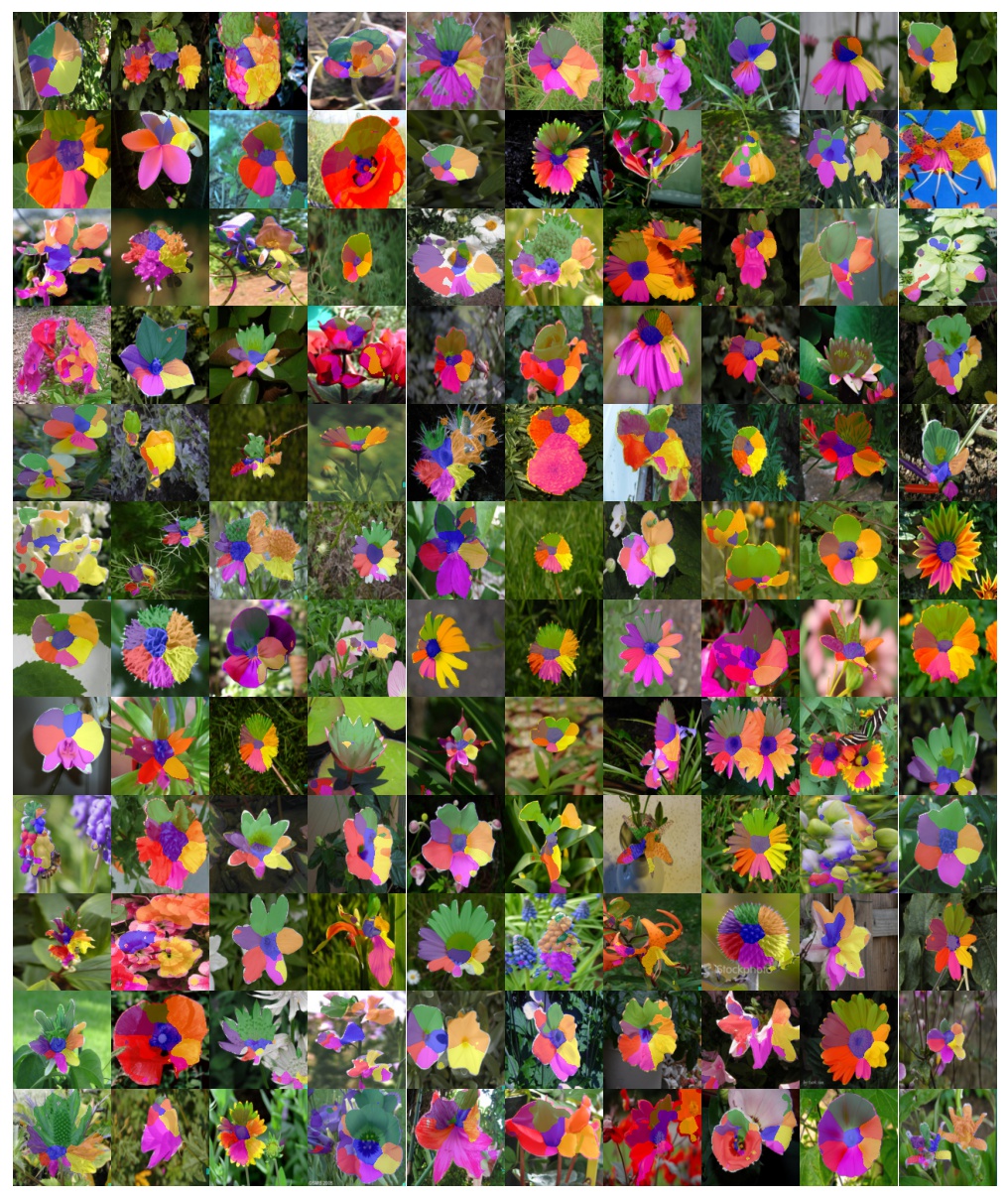}
\end{center}
   \caption{\textbf{120 samples from Flower,} with detected masks overlaid.}
   \label{fig:det_flower}
\end{figure*}

To support the representative nature of the results shown in the main paper, we sample hundreds of images at random to evaluate both the quality of our generator as well as the keypoint detector applied to real images. Figures~\ref{fig:gen_celeba}, \ref{fig:gen_taichi}, \ref{fig:gen_cub}, and \ref{fig:gen_flower} show the randomly generated images along with the corresponding points and masks. Figures~\ref{fig:det_celeba}, \ref{fig:det_taichi}, \ref{fig:det_cub}, and \ref{fig:det_flower} show the detected masks on real test images sampled at random.

\section{Implementation Details}

\subsection{Integrated Network Architecture} \label{sec:int_archi}
\begin{figure*}[t]
\begin{center}
   \includegraphics[width=0.98\linewidth]{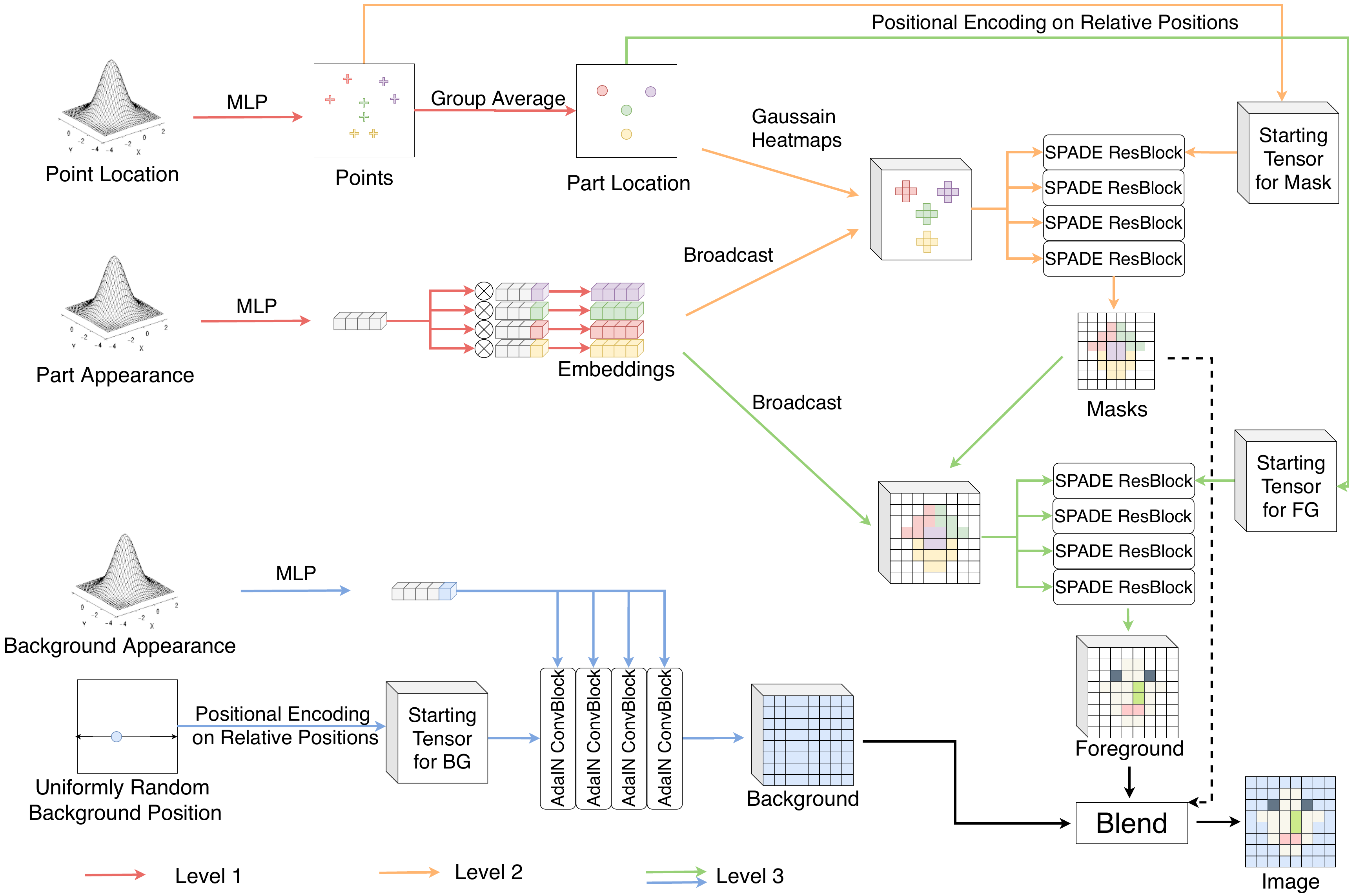}
\end{center}
   \caption{\textbf{Integrated network architecture} for better understanding of our data flow.}
\label{fig:int_archi}
\end{figure*}

We provide a complete overview of the network architecture in Figure~\ref{fig:int_archi}.
It combines the individual modules introduced in the main paper.

\subsection{SPADE ResBlock} \label{sec:spade}
Figure~\ref{fig:spade} (top) illustrates the SPADE ResBlock \cite{park2019semantic}. To be self-contained, we summarize the implementation of SPADE here. SPADE takes two feature maps as input. We denote the two feature maps as $\mF_\text{input}\in\R^{D_\text{input\_emb}\times H\times W}$ and $\mF_\text{style}\in\R^{D_\text{style\_emb}\times H\times W}$. We first use BatchNorm \cite{ioffe2015batch} to normalize $\mF_\text{input}$ followed by two convolutions to map $\mF_\text{style}$ to the new mean $\boldsymbol \beta\in\R^{D_\text{input\_emb}\times H\times W}$ and new standard deviation $\boldsymbol \gamma\in\R^{D_\text{input\_emb}\times H\times W}$ for the normalized $\mF_\text{input}$. Although we use BatchNorm to normalize $\mF_\text{input}$ in batch and channel dimension, we apply the generated $\boldsymbol \beta$ and $\boldsymbol \gamma$ to denormalize every individual element.

\subsection{AdaIN ConvBlock} \label{sec:adain}
\begin{figure}[t]
\begin{center}
   \includegraphics[width=0.98\linewidth]{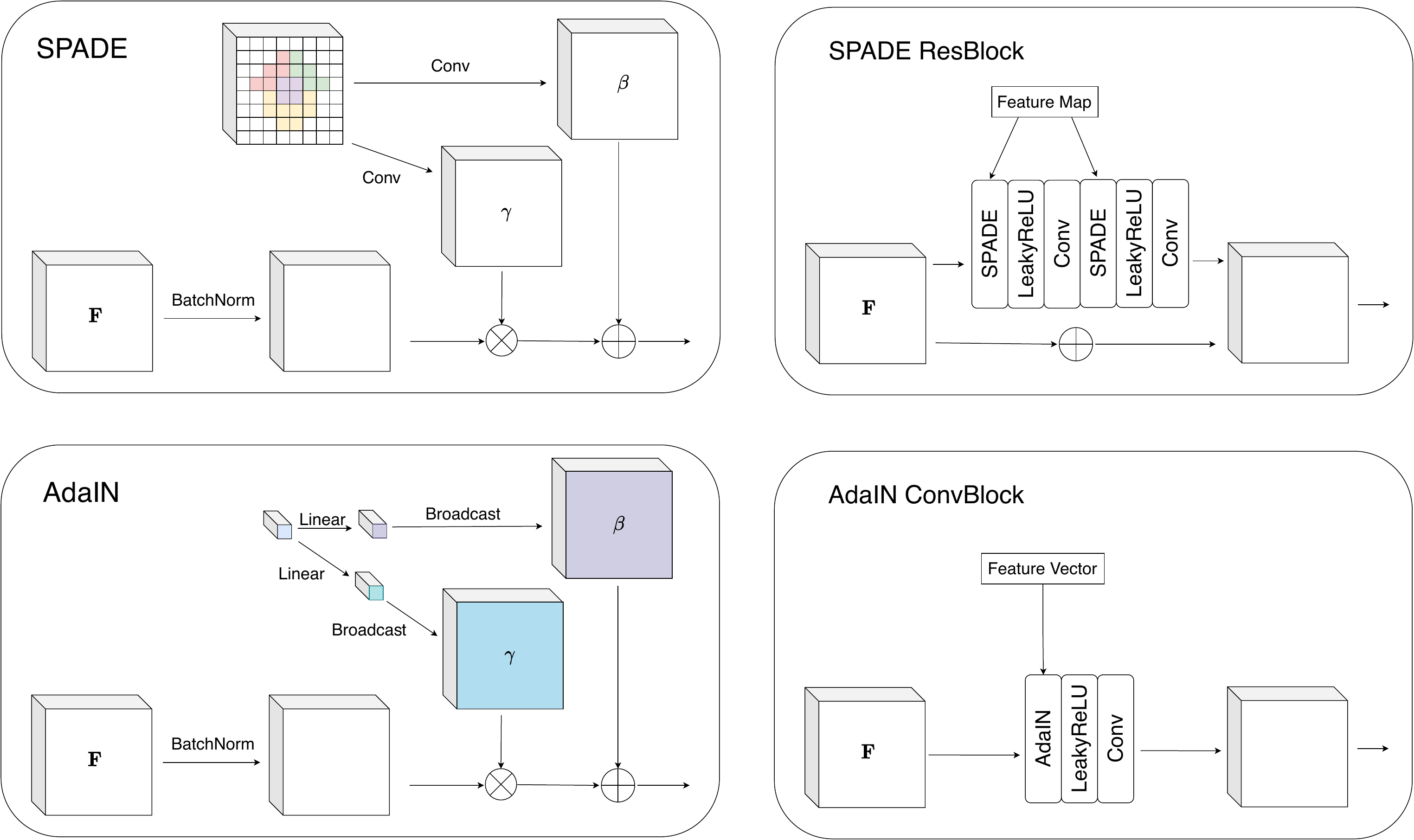}
\end{center}
   \caption{\textbf{SPADE ResBlock.} (top) and \textbf{AdaIN ConvBlock} (bottom).}
\label{fig:spade}
\end{figure}

Figure~\ref{fig:spade} (bottom) illustrates the AdaIN ConvBlock \cite{huang2017arbitrary}. AdaIN takes a feature map and a style vector as input. We denote the feature map as $\mF_\text{input}\in\R^{D_\text{input\_emb}\times H\times W}$ and the style vector as  $\vv_\text{style}\in\R^{D_\text{style\_emb}}$. Unlike the original paper \cite{huang2017arbitrary}, which uses InstanceNorm \cite{ulyanov2016instance}, we found it beneficial in our case to use BatchNorm \cite{ioffe2015batch} to normalize $\mF_\text{input}$. We use two fully connected layers to map $\vv_\text{style}$ to the new mean $\boldsymbol \beta\in\R^{D_\text{input\_emb}}$ and new standard deviation $\boldsymbol \gamma\in\R^{D_\text{input\_emb}}$ for the normalized $\mF_\text{input}$. As for the SPADE block, we use BatchNorm to normalize $\mF_\text{input}$ in batch and channel dimension. For denormalization, $\boldsymbol \beta$ and $\boldsymbol \gamma$ are generated for each channel and broadcast to the spatial size of the feature map.

\subsection{Hyperparameters}
We use the Adam optimizer \cite{KingmaB14} with a learning rate of 0.0001 for the generator $\cG$ and 0.0004 for the discriminator $\cD$, both with $\beta_1=0.5$, $\beta_2=0.9$. We set the gradient penalty coefficient $\lambda_\text{gp}=10$ for the discriminator. In every experiment, we update the generator 30,000 times. The learning rate for DeepLab is 0.0003, with $\beta_1=0.9$, $\beta_2=0.999$. DeepLab is trained for 10,000 iterations. The feature map resolutions are ($32^2,32^2,64^2,128^2$), respectively. 
We set $n_\text{per}=4$ for all experiments. The penalty coefficients for each experiment is listed in Table~\ref{tab:penalty_coef}. For Taichi, we set the vertical position of the background to be always in the center and only sample the horizontal position uniformly.

\begin{table}[ht]
\centering
\begin{tabular}{|c|c|c|c|c|}
\hline
& CelebA & Taichi & CUB & Flower  \\ \hline
$\lambda_\text{con}$ & 10 & 30 & 10 & 30 \\ \hline
$\lambda_\text{area}$ & 1 & 1 & 1 & 1 \\ \hline
\end{tabular}
\caption{\textbf{Penalty coefficients} used in the main paper.} 
\label{tab:penalty_coef}
\end{table}
\section{Image Quality}
To quantify image quality, we report the FID score \cite{Seitzer2020FID} for the models used in the main paper: 21.42 for CelebA-in-the-wild, 116.62 for Taichi, 38.74 for CUB, and 38.41 for Flowers. Note that FID numbers are not comparable across datasets. The FID is calculated by 50,000 generated images and the corresponding original dataset as established by \cite{karras2019style}. Note that image quality is not our primary goal but it is still important since it influences how well the learned detector will generalize. The better the quality, the smaller the domain gap between generated and real images.

\section{Embedding Visualization}
\begin{figure}[ht]
\begin{center}
   \includegraphics[width=0.98\linewidth]{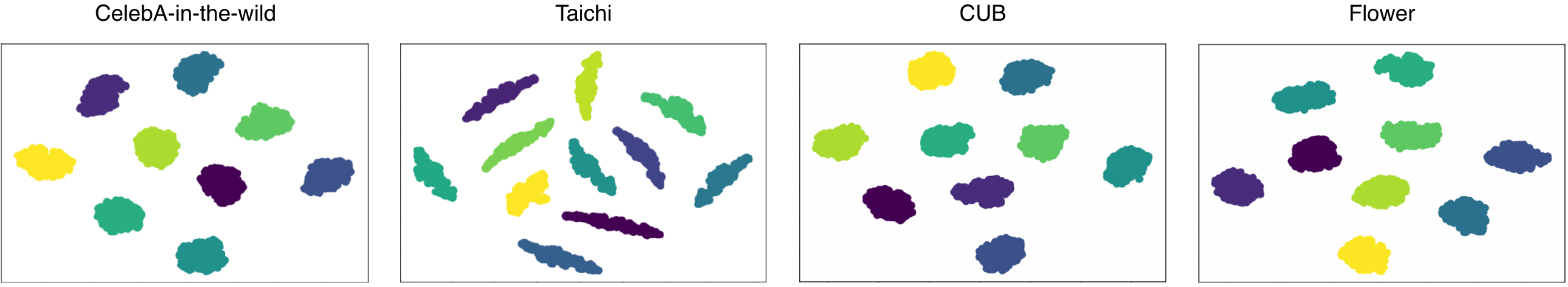}
\end{center}
   \caption{\textbf{Embedding visualization.}}
\label{fig:emb_vis}
\end{figure}

We generated 320 images for each dataset and provide the T-SNE visualization of their embedding vectors in Figure~\ref{fig:emb_vis}, including the background as a part. The embeddings from different parts are well separated in all datasets, indicating that our model learns unique embeddings for each part. Since overlapping part embeddings would mean that the same image features could be generated by multiple parts, the separation explains the good consistency of parts across images. 

\section{Theoretical Analysis} \label{sec:theory}

In the following, we explain in detail how and why the proposed GAN architecture provides translation equivariance.
On top of the translation equivariance of convolutions, we address equivariances in the point- and part-conditioned  image generation.

\begin{figure}[ht]
\begin{center}
   \includegraphics[width=0.98\linewidth]{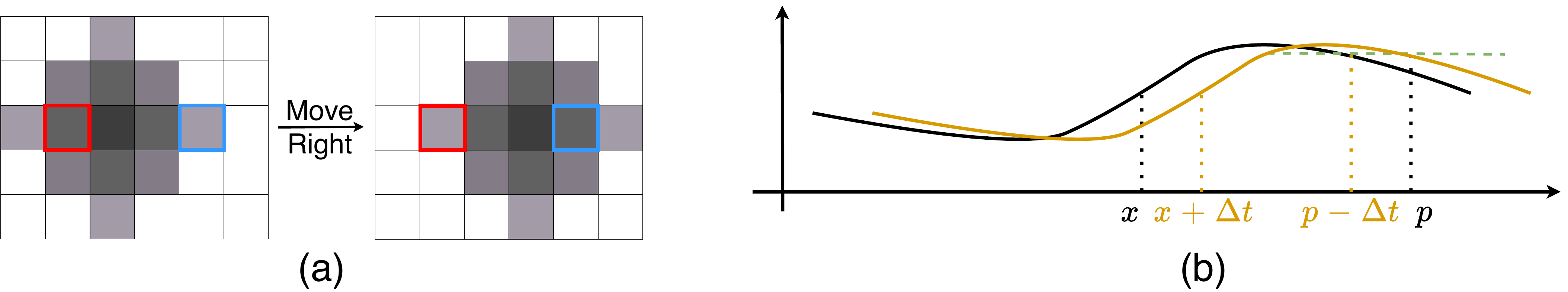}
\end{center}
   \caption{\textbf{Intuition.} (a) We move the points to the right by one pixel. The left part of the point (red box) becomes brighter and the right part of the point (blue box) becomes darker. (b) Continuous 1D case. (c) Rotation of parts can be achieved by translating points.}
\label{fig:intuition}
\end{figure}

\paragraph{Intuition.} Consider a greyscale image, with a single point on it. Assume the image gets darker to the point. How should the image change if we move the point one pixel step to the right? We illustrate it in Figure~\ref{fig:intuition} (Left). The left part of the image w.r.t. the point gets brighter and the right part gets darker. More specifically, if the point moves one pixel to the right, the pixels on the grid take the value of their left neighbor.
The continuous case (1D for simplicity) is shown in Figure~\ref{fig:intuition} (Middle). We assume that the point $x$ controls the translation of the function $f$. If we move the $x$ to $x+\Delta t$, the function $f$ is also shifted by $\Delta t$. Then for each fixed position $p$, the value changes from $f(p)$ to $f(p-\Delta t)$. We desire a network architecture that fulfills this equivariance in the vicinity of each point.

\paragraph{Formulation.} Formally, let $\mI\in\R^{H\times W}$ denote a feature map (or an image), $\vx\in[-1,1]\times [-1, 1]$ denote the point location, and $\vp\in[-1, 1]\times [-1, 1]$ denote the pixel location. We have
\begin{equation}
    \mI(\vp-\Delta \vt, \vx) = \mI(\vp, \vx+\Delta t).
\label{eq:single_kp_motion_initial}
\end{equation}
If we divide both sides by $\Delta \vt$ (both axis independently), and let $\Delta \vt \rightarrow \mathbf 0$, we get the equation for the motion of a single point
\begin{equation}
    -\frac{\partial \mI(\vp, \vx)}{\partial \vp} = \frac{\partial \mI(\vp, \vx)}{\partial \vx}.
\label{eq:single_kp_motion}
\end{equation}
This equation is the first integral of a homogeneous linear partial differential equation \cite{xiong2017boundary, xiong2019identifying}. Its solution is in the form of $\mI(\vp - \vx)$, where $\mI$ can be any function. In other words, whenever $\vp$ (or $\vx$) shows up, it must be in a combination with $\vx$ (or $\vp$) as $\vp - \vx$ or $\vx - \vp$.

If there exist multiple points, each pixel may be influenced by multiple points. Assuming $K$ points $\vx_1,...,\vx_K\in[0,1]^2$ gives
\begin{equation}
    \mI_k(\vp-\Delta \vt, \vx_1,...,\vx_K) = \mI_k(\vp, \vx_1,..., \vx_k+\Delta t, ...,\vx_K),
\label{eq:multi_kp_motion_initial}
\end{equation}
where $k=1,...,K$.

We assume additivity of $\mI_k$, which models the additivity of the generated feature maps in our convolutional generator well,
\begin{equation}
    \mI = \sum_{k=1}^K\mI_k,
\label{eq:additivity_assumption}
\end{equation}
and write
\begin{equation}
     \mI(\vp-\Delta \vt, \vx_1,...,\vx_K) = \sum_{k=1}^K\mI_k(\vp, \vx_1,..., \vx_k+\Delta t, ...,\vx_K).
\label{eq:multi_kp_motion_initial_additivity}
\end{equation}
As before, we divide both sides by $\Delta \vt$ (both axis independently), and let $\Delta \vt \rightarrow \mathbf 0$. We get
\begin{equation}
    \frac{\partial \mI(\vp, \mX)}{\partial \vp} = - \sum_{k=1}^K \frac{\partial \mI_k(\vp, \mX)}{\partial \vx_k}.
\label{eq:multi_kp_motion}
\end{equation}
where $\mX=\{\vx_1,...,\vx_K\}$ for simplicity. Similarly, the solution to this equation is in the form of $\mI(\vp - \vx_1, ..., \vp - \vx_k)$. Therefore, in theory, if we build a neural network that takes the difference between the points and the pixels as input, i.e., $(\vp - \vx_1, ..., \vp - \vx_k)$, this equation is automatically satisfied. Hence, unlike recent trends of adding absolute coordinates to networks \cite{islam2020much, xu2020positional, dosovitskiy2020image, chu2021conditional}, we need to remove the absolute coordinate and point information from the model.

\paragraph{Background Handling.} We generate foreground and background independently as foreground and background are not additive. The background exists on all pixels of the image, just with parts occluded by the foreground. Thus the movement of the background will not affect the foreground pixels.  Mathematically, if only a non-empty subset of the $K$ points influence the pixel $\vp$, Equation~\ref{eq:multi_kp_motion} still holds because $\frac{\partial \mI(\vp, \vx_1,..., \vx_K)}{\partial \vx_k}=0$ if $\vp$ is not influenced by the point $\vx_k$. However, if $\vp$ belongs to the background (no point affects the pixel $\vp$), the RHS becomes zero while the LHS does not (unless the background has only a single color). Therefore, we need to separate foreground and background and blend them at the end, to be able to assume additivity only on the foreground where it is meaningful.

\paragraph{Convolution.} We use convolutions in our network since they are translation equivariant. In formulation, a neighbor of $\vp$ has value $\mI(\vp - \vx_1 +\Delta \vp, ..., \vp - \vx_k+\Delta \vp)$ where $\Delta\vp$ is a step to the neighbor. Note that $\mI(\vp - \vx_1 +\Delta \vp, ..., \vp - \vx_k + \Delta \vp)$ itself can be written as a function of $(\vp - \vx_1, ..., \vp - \vx_K)$, which makes our desired Equation~\ref{eq:multi_kp_motion} hold. 

\comment{
\paragraph{Transformation.} We have demonstrated that we can use the form of $\mI(\vp-\vx)$ to achieve translation equivariance of points. We argue that the translation equivariance of points can be used to infer the general 2D transformation of parts and objects. As shown in Figure~\ref{fig:intuition} (c), the rotation of a part can be achieved by the translation of the points.  Hence, we can translate several points to achieve the rotation of a part. The same idea applies for general 2D transformations.}

\paragraph{Why to use a GAN?} Unlike most previous work which uses auto-encoders, we first use a GAN to generate points, masks, and images, and train a segmentation network to obtain segments. A core reason for us is that GANs can effectively prevent leaking absolute position \cite{karras2021alias}. As pointed out by \cite{islam2020much, alsallakh2021mind, xu2020positional, kayhan2020translation}, a convolution (kernel size larger than 1) with zero padding implicitly encodes absolute grid position. The more layers and downsampling layers, the larger is the region around the boundary that is affected by leaking absolute position \cite{alsallakh2021mind}. However, if we were to remove the zero-padding, using only valid convolution, encoder will perform much worse at the boundary of images \cite{murase2020can}. GANs, however, usually do not have downsampling but multiple upsampling layers. If we maintain a fixed margin around the feature map and crop after each upsampling \cite{karras2021alias}, we can effectively prevent leaking absolute position and work with cropped and uncropped datasets.
Moreover, autoencoders require to learn the encoder and decoder together, which implies a larger memory footprint and they generally lack behind in image generation quality.
\section{Additional Ablation Tests}

\begin{figure*}[t]
\begin{center}
   \includegraphics[width=0.9\linewidth]{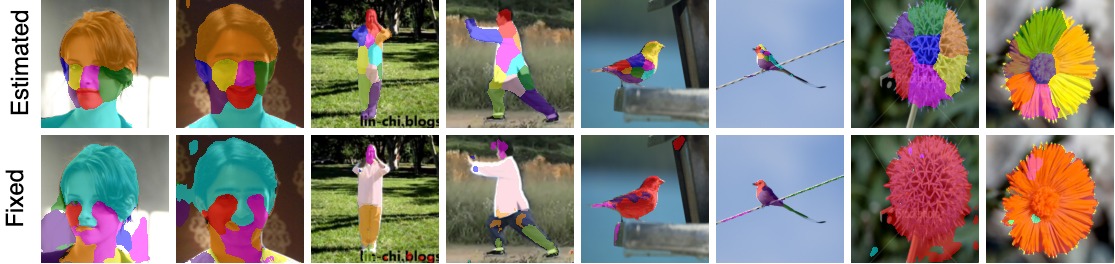}
\end{center}
\vspace{-18pt}
   \caption{\textbf{Ablation Test on Estimated $\sigma$ versus Fixed $\sigma$}. Examples are given for four different datasets.}
   \label{fig:ablation_fixed}
\end{figure*}

\subsection{Replacing the GAN with an Auto-encoder}
\begin{figure}[t]
\begin{center}
   \includegraphics[width=0.98\linewidth]{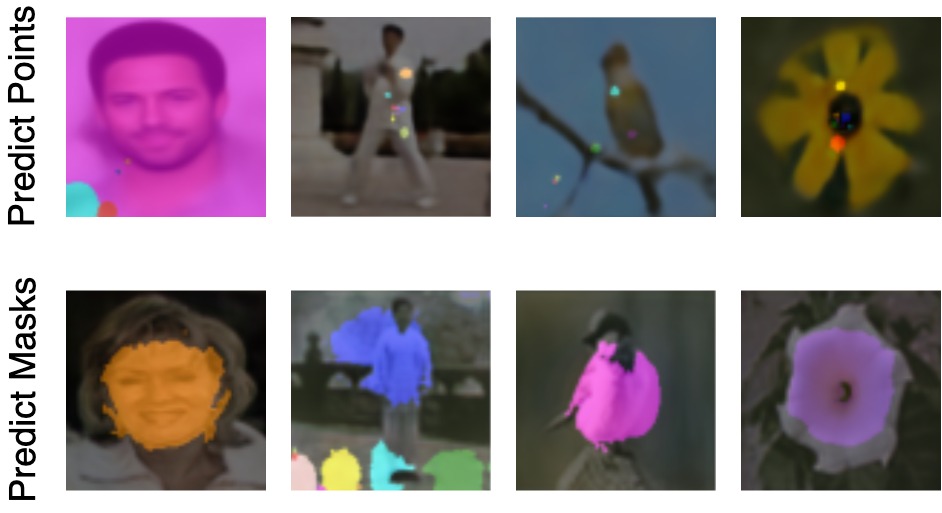}
\end{center}
\vspace{-15pt}
   \caption{\textbf{Ablation Test on replacing the GAN with an auto-encoder.} Detected points and masks are overlayed on the corresponding images. In this experiment, points move to degenerate positions and scales. Masks are inaccurate.}
   \label{fig:autoencoder}
\end{figure}

We tried two kinds of auto-encoders to replace our Point Generator (Level 1 of the hierarchy) and Mask Generator (Level 2). In the first one, we train a U-Net \cite{ronneberger2015u} to obtain points $\{\vx_k^{1}, ... \vx_k^{n_\text{per}}\}_{k=1}^K$, and two separate ResNet-18 \cite{he2016deep} to extract part appearance vector $\vw^\text{dynamic}$, and background appearance vector $\vw_\text{bg}$ along with background position $\vu_\text{bg\_pos}$, to feed into our Mask Generator that is left unchanged. In the second version, we directly use DeepLab V3 to generate masks $\mM$, and then feed it to our Foreground and Background generator. We use the mean squared image reconstruction error to train this auto-encoder. We observe that all of the tested auto-encoders give trivial solutions, as shown in Figure~\ref{fig:autoencoder}. Hence, we decided to use a pure GAN setup.

\subsection{Using Fixed Standard Deviation} 
Instead of estimating from points, we directly set a fixed standard deviation $\sigma$ for each part. The fixed standard deviation is calculated by the average of the $\sigma_k$ from our pre-trained model. We sample 5000 images. The average for all datasets are 0.00725 (CelebA-in-the-wild), 0.010 (Taichi), 0.0016 (CUB), and 0.0065 (Flower). The quantitative results are shown in Table~\ref{tab:supp_ablation_fixed} and the qualitative results are shown in Figure~\ref{fig:ablation_fixed}. The results show that fixing $\sigma$ harms the performance severely. It is important to have various $\sigma$ for each part and each object.

\begin{table}[t]
\centering
\resizebox{0.98\linewidth}{!}{%
\begin{tabular}{|l|c|c|c|c|c|c|}
\hline
Method & CelebA $\downarrow$ & CUB $\uparrow$ & Flowers $\uparrow$ & Taichi (MAE) $\downarrow$ & Taichi (IoU) $\uparrow$ \\ \hline
fixed $\sigma$ & 26.32\%  & 0.467 & 0.697 & 657.61 & 0.7452 \\
original & \textbf{6.18}\%  & \textbf{0.629} & \textbf{0.739} & \textbf{417.17} & \textbf{0.8538} \\ \hline
\end{tabular}}
\vspace{-5pt}
\caption{\textbf{Quantitative Ablation Tests on Number of Parts}. The metric for each dataset in this table follows the main paper.}
\label{tab:supp_ablation_fixed}
\end{table}

\subsection{Number of Points per Part}
We test on the influence of the number of points per part $n_\text{per}$. We found that the network always collapses if $n_\text{per}\leq 2$, and collapses occasionally if $n_\text{per}=3$. Table~\ref{tab:supp_ablation_test_nper} shows that the optimal $n_\text{per}$ differs among different datasets. We choose $n_\text{per}=4$ in our main paper for consistency.

\begin{table}[t]
\centering
\resizebox{0.98\linewidth}{!}{%
\begin{tabular}{|c|c|c|c|c|c|c|}
\hline
$n_\text{per}$ & CelebA $\downarrow$ & CUB $\uparrow$ & Flowers $\uparrow$ & Taichi (MAE) $\downarrow$ & Taichi (IoU) $\uparrow$ \\ \hline
3 & \textbf{6.08}\%  & 0.644 & 0.696 & 594.73 & 0.5863 \\
4 & 6.18\%  & 0.629 & 0.739 & \textbf{417.17} &\textbf{0.8538} \\
6 & 8.84\%  & \textbf{0.682} & 0.715 & 434.47 & 0.8336 \\
8 & 13.02\% & 0.641 & \textbf{0.744} & 467.69 & 0.7792 \\\hline
\end{tabular}}
\vspace{-5pt}
\caption{\textbf{Ablation Tests on Number of Points per Part}. The number of parts and the metric for each dataset in this table follow the main paper. We use $n_\text{per}=4$ in the main paper.}
\label{tab:supp_ablation_test_nper}
\end{table}

\subsection{Number of Parts}
We test how the number of parts affects segmentation. Although the Area Loss is vital, its coefficient 
is not sensitive to the 
 numbers of parts $K$. Thus for simplicity, we set $\lambda_\text{area}=1$ as in the main paper. However, we found that small number of parts with large Concentration Loss coefficient causes the trivial or sub-optimal solutions. Therefore, we use the formula 
\begin{equation}
    \lambda_\text{con}(K) = C_\text{con} K
\end{equation}
The constant $C_\text{con}$ varies between datasets. In our experiments, we set 1.25 for Celeba-in-the-wild, 3 for Taichi, 1.25 for CUB, and 3.75 for Flower. 

\begin{figure*}[t]
\begin{center}
   \includegraphics[width=0.98\linewidth]{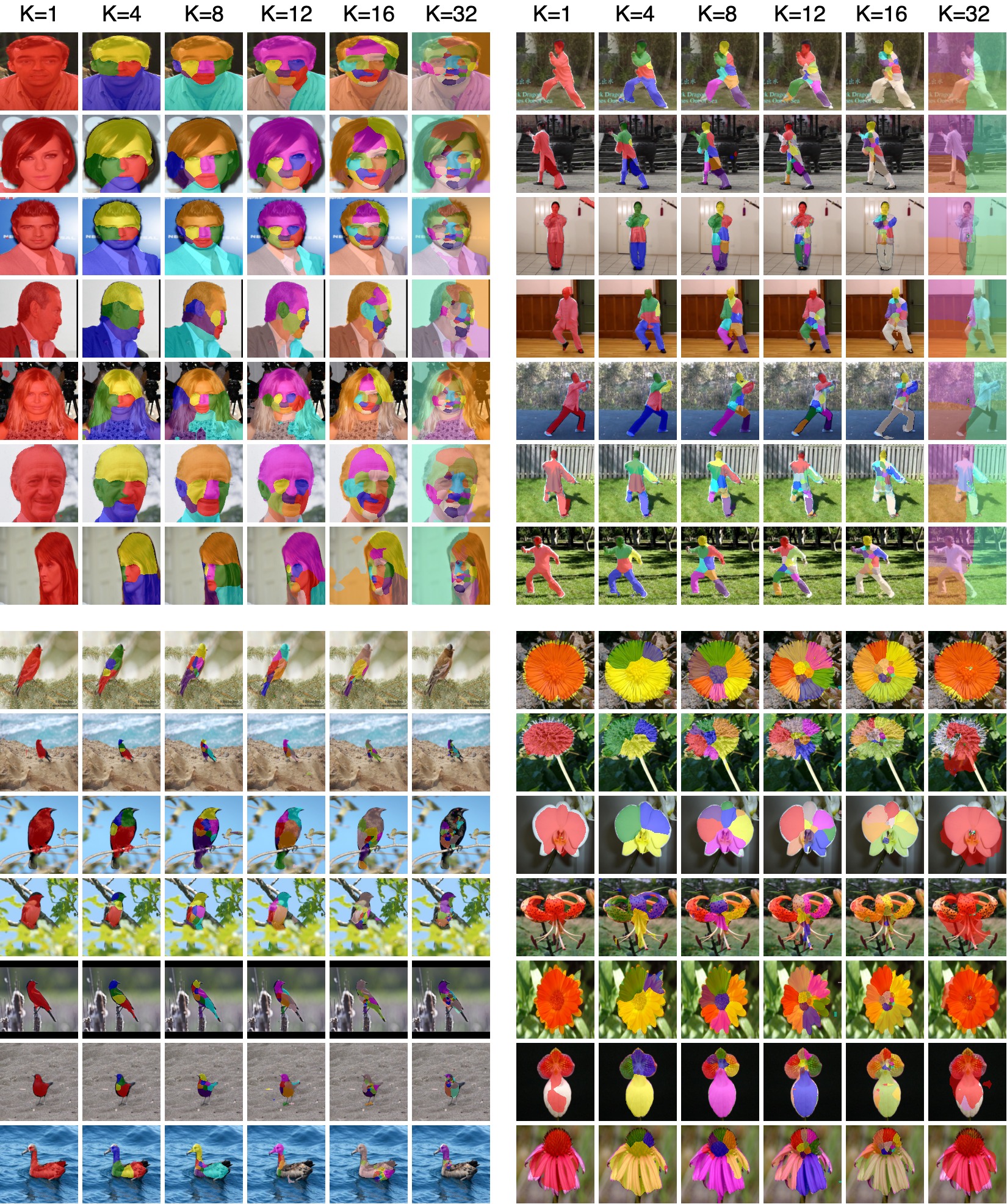}
\end{center}
   \caption{\textbf{Ablation Test on the number of parts.} Seven examples are tested for each dataset.}
   \label{fig:ablation_n_parts}
\end{figure*}

We show quantitative results in Table~\ref{tab:supp_ablation_test_n_parts} and qualitative results in Figure~\ref{fig:ablation_n_parts}. We found our model starts to degenerate when $K=16$. Even though the IoU on CUB is higher, the mask degenerates, which is still a state-of-the-art foreground/background segmentation model but not a good part model. If $K=32$, the model fails to distinguish the foreground from the background.

\begin{table}[t]
\centering
\resizebox{0.98\linewidth}{!}{%
\begin{tabular}{|l|c|c|c|c|c|c|}
\hline
Method & CelebA $\downarrow$ & CUB $\uparrow$ & Flowers $\uparrow$ & Taichi (MAE) $\downarrow$ & Taichi (IoU) $\uparrow$ \\ \hline
K=1 & 56.23\%  & 0.631 & 0.570 & 748.30 & 0.7895 \\
K=4 & 12.26\%  & 0.607 & \textbf{0.767} & 652.87 & 0.8176 \\
K=8 & 6.18\%  & 0.629 & 0.739 & \textbf{420.95} & \textbf{0.8481} \\
K=12 & \textbf{6.17}\%  & 0.663 & 0.708 & 450.51 & 0.7939 \\
K=16 & 6.71\%  & \textbf{0.692} & 0.712 & 441.04 & 0.7475 \\
K=32 & 7.00\%  & 0.380& 0.530 & 1605.84 & 0.1758 \\\hline
\end{tabular}}
\vspace{-5pt}
\caption{\textbf{Quantitative Ablation Tests on Number of Parts}. The metric for each dataset in this table follows the main paper.}
\label{tab:supp_ablation_test_n_parts}
\end{table}

\end{document}